\documentclass[journal]{IEEEtran}
\IEEEoverridecommandlockouts
\usepackage{float}

\usepackage{mathtools}
\usepackage{amsthm, amssymb}
\usepackage{mathrsfs}
\usepackage{hyperref}
\usepackage{enumitem}
\usepackage{subcaption}
\usepackage{makecell}
\usepackage{svg}
\usepackage{multirow}
\definecolor{revision}{rgb}{0.0, 0.2, 0.8}
\usepackage{algorithm}
\usepackage{algpseudocode}
\usepackage{array}
\usepackage{booktabs}
\usepackage{makecell}
\usepackage{graphicx}
\newcolumntype{L}[1]{>{\raggedright\arraybackslash}p{#1}}

\theoremstyle{plain}
\newtheorem{theorem}{Theorem}
\newtheorem{lemma}{Lemma}
\newtheorem{assumption}{Assumption}
\newtheorem{remark}{Remark}
\newtheorem{prop}{Proposition}
\newtheorem{definition}{Definition}

\newtheorem{condition}{Condition}
\newenvironment{proofsketch}{\begin{IEEEproof}[Proof sketch]}{\end{IEEEproof}}

\usepackage{microtype}
\SetExpansion[context=bibliography,stretch=30,shrink=30,step=5]
  {encoding={OT1,T1,TS1}}{}
\usepackage{cite}
\usepackage{amsmath,amssymb,amsfonts}
\usepackage{graphicx}
\usepackage{textcomp}
\usepackage{xcolor}
\def\BibTeX{{\rm B\kern-.05em{\sc i\kern-.025em b}\kern-.08em
    T\kern-.1667em\lower.7ex\hbox{E}\kern-.125emX}}

\begin{document}

\newcommand{\bo}[1]{{\footnotesize\color{red}[Bo: #1]}}
\def\yin{\color{blue}Yin:}

\title{Multimodal Remote Inference
{
}

\thanks{This work was supported in part by the National Science Foundation (NSF) under Grants CNS-2106427, CNS-2312833, CNS-2239677, and by the Alabama Department of Economic and Community Affairs under Grant No. 1ARDEF2611. A preliminary version of this work focusing on the two-modality case was presented at IEEE MASS 2025~\cite{11206222}.}
}

\author{\IEEEauthorblockN{Keyuan Zhang\IEEEauthorrefmark{1}, Yin Sun\IEEEauthorrefmark{2}, Bo Ji\IEEEauthorrefmark{1}}

\IEEEauthorblockA{\IEEEauthorrefmark{1}Department of Computer Science, Virginia Tech \\ \IEEEauthorrefmark{2}Department of Electrical and Computer Engineering, Auburn University}

\IEEEauthorblockA{Email: \{keyuanz, boji\}@vt.edu, yzs0078@auburn.edu}
}

\maketitle

\begin{abstract}
We consider a remote inference system with multiple modalities, where a multimodal machine learning (ML) model performs real-time inference using features collected from remote sensors.
When sensor observations evolve dynamically over time, fresh features are critical for inference tasks.
However, timely delivery of features from all modalities is often infeasible under limited network resources.
To address this challenge, we formulate a multimodal scheduling problem to minimize the ML model’s inference error. We model this error as a general function of the Age of Information (AoI) vector, where AoI quantifies data freshness.
We cast the problem as a semi-Markov decision process (SMDP) and derive an equivalent reformulation with a reduced state set.
We then show that the problem has fundamentally different chain structures in the two-modality and multi-modality cases.
For the two-modality case, we prove that the optimal policy has an index-based threshold structure.
{For the general multi-modality case (two or more modalities), we develop an optimal switching-type policy (OPT-ST), which is computed using a multichain policy iteration algorithm (MPI). To further reduce complexity, we also develop two low-complexity policies: the round-robin-like switching-type policy (RR-ST) and the round-robin-like threshold policy (RR-Thres).}
{Numerical results show that OPT-ST achieves the lowest inference error among all baselines, with reductions of up to 80.3\%. In a five-modality case, RR-ST and RR-Thres achieve $5.2\times$ and $89.8\times$ speedups in offline computation over OPT-ST, respectively, at the cost of increasing the inference error by 23.3\% and 38.6\%, respectively.}
\end{abstract}


\section{Introduction}







\begin{figure}
    \centering
    \includegraphics[width=0.90\linewidth]{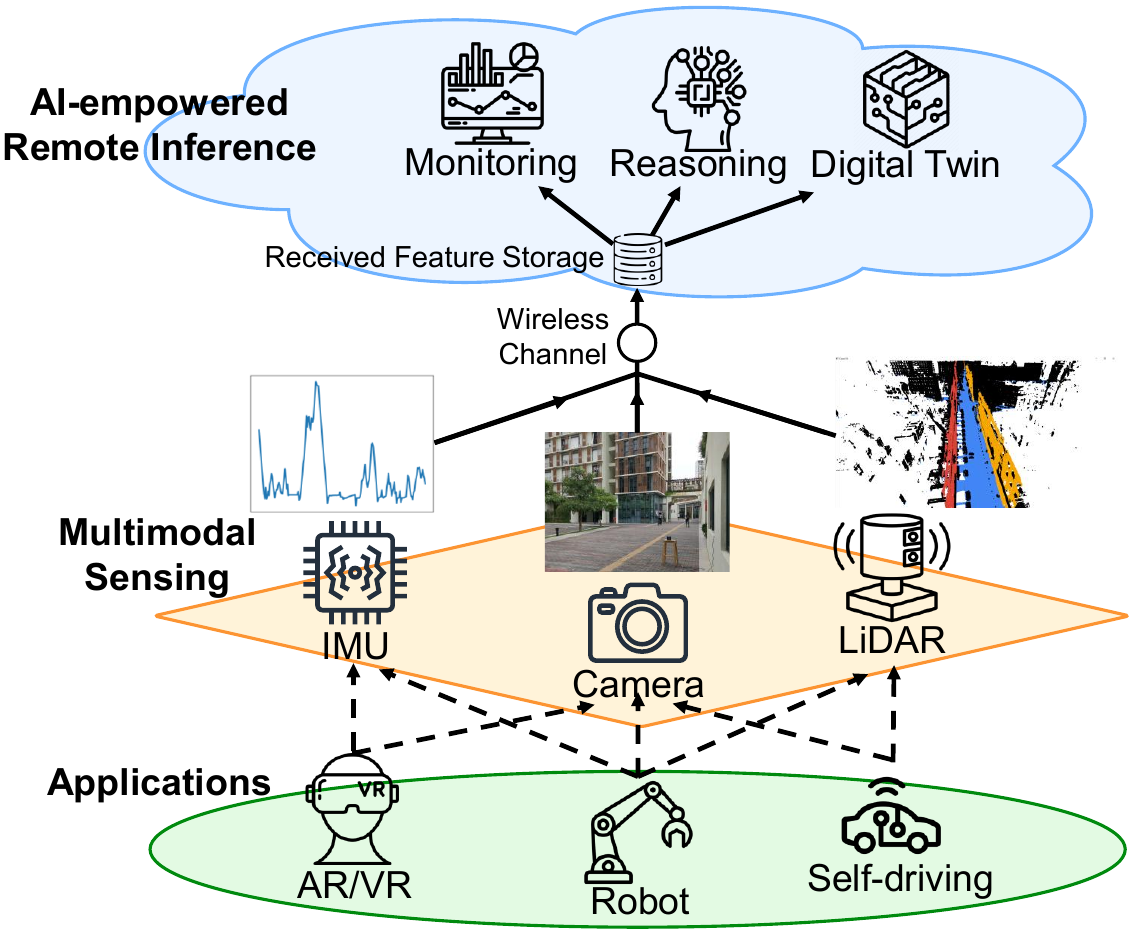}
    \caption{A multimodal remote inference system. 
    }
    \label{fig:application}
\end{figure}

The advent of sixth-generation (6G) technology, along with advances in artificial intelligence (AI), enables \emph{remote inference}~\cite{10559951} in various intelligent applications, such as autonomous transportation, unmanned mobility, and industrial automation.
As illustrated in Fig.~\ref{fig:application}, a remote inference system uses AI models for inference tasks (e.g., monitoring, reasoning, and decision-making) based on features transmitted from remote sensors.
For instance, traffic prediction relies on near real-time forecasts of traffic status (e.g., speed, flow, and 
demand) based on spatio-temporal road data~\cite{yuan2021survey}.
When the sensor observations change dynamically over time, timely data delivery is critical.
For example, autonomous driving requires timely updates on vehicle positions and velocities to ensure safety; healthcare monitoring relies on real-time vital signs for timely alerts. 

Moreover, such complex tasks often involve multiple data modalities, such as audio, visual, 3D points (e.g., LiDAR or RADAR), and motion (e.g., IMU) data.
Each modality provides complementary information to enhance the overall inference accuracy.
For example, in object detection and tracking, RGB images capture object shape and appearance, while LiDAR provides depth information~\cite{7795718}.
To exploit such information, machine learning (ML) methods such as deep neural networks are widely used to extract modality-specific features and capture cross-modal correlations~\cite{10.1145/3656580}.

Despite the advantages of multimodal ML, its application in remote inference systems remains limited in practice.
One main challenge is that multimodal ML models often require features from different modalities to be aligned and synchronized~\cite{jiang2025multi}.
However, under limited network resources, such as bandwidth or channel constraints, it is often difficult to deliver fresh features from all modalities simultaneously.
Because the freshness of each modality evolves dynamically over time, the scheduler needs to decide which modality to schedule to improve the ML model’s performance.
Moreover, due to the complexity of multimodal ML models, finding a modality scheduling policy to minimize inference error can be difficult.

To address this challenge, prior work on remote inference has provided important insights into optimizing ML inference performance, which is expressed as a function of Age of Information (AoI)~\cite{8000687, 8764465, 9484640, shisher2022does, 10559951, ari2026goal, shisher2025aoi}.
Age of Information, introduced in~\cite{6195689}, is an important metric for information freshness and is defined as the time elapsed since the freshest received information was generated.
Most existing studies focus on either the single-source setting or multi-source settings with independent sources~\cite{8000687, 8764465, 9484640, shisher2022does, 10559951, ari2026goal}, where the objective function is a weighted sum of scalar functions of AoI.
In multimodal remote inference, however, the inference error is generally an AoI vector function, denoted by \(L(\boldsymbol{\Delta}(t))\), where \(\boldsymbol{\Delta}(t)=(\Delta_1(t),\dots,\Delta_M(t))\) is the AoI vector of the \(M\) modalities at time \(t\).
This function can be quite general: it may be non-monotonic, non-convex, and non-separable, which further complicates the scheduling problem.
To this end, in this paper, we study the following key research question:

\emph{How to design an effective and efficient modality scheduling policy to minimize multimodal remote inference error?} 

We answer this question by making the following main contributions:

\begin{itemize}
\item
We formulate a multi-modality scheduling problem that minimizes the inference error, which is a function of the AoIs across modalities.
We cast it as a semi-Markov decision process (SMDP), and classical algorithms such as policy iteration and value iteration are computationally expensive due to the curse of dimensionality.
To mitigate the issue, we derive an equivalent reformulation that reduces the state set size from at most \(K^M\) to at most \(M(M-1)K^{M-2}\), where \(M\) denotes the number of modalities and \(K\) the maximum AoI.
In typical multimodal settings, the number of modalities is relatively small (\(M<K\)). The reformulation therefore simplifies computation by reducing the exponent from \(M\) to \(M-2\) while preserving optimality.
For \(M\!=\!4\) and \(K\!=\!50\), the reduction factor is about \(208\).
\item
For the two-modality case ($M\!=\!2$), we derive a closed-form solution for the optimal policy. 
{
Specifically, we derive a closed-form solution to the Bellman optimality equation, which reveals an \emph{index-based threshold structure}.}
Under this policy, one modality is scheduled continuously until its index reaches a threshold, at which point the scheduler switches to the other modality. Interestingly, the two modalities share the same threshold.
\item
{For the general case (\(M\ge2\)), we develop an optimal switching-type policy (OPT-ST)}: the scheduler jointly determines the number of consecutive transmissions of the current modality and the next modality to switch to. {When $M>2$, the problem may be \emph{multichain}, rendering standard unichain policy iteration inapplicable. We therefore use \emph{multichain policy iteration} (MPI), which applies to both unichain and multichain problems, to compute OPT-ST.}

\item
{To further reduce complexity for $M>2$, we consider two progressively more restricted policy classes. The resulting algorithms also apply when $M=2$.} First, we consider a round-robin-like switching-type policy (RR-ST) with a fixed switching order, which can also be computed via MPI. Second, we consider a round-robin-like threshold policy (RR-Thres), where each modality has a single threshold; the thresholds are computed via coordinate descent. {With transition data precomputed, the per-iteration time complexities of OPT-ST and RR-ST are \(\mathcal{O}\!\left(M(M-1)^2K^{M-1}\right)\) and \(\mathcal{O}\!\left(MK^{M-1}\right)\), respectively. RR-Thres has a per-iteration time complexity of \(\mathcal{O}(MK)\), as detailed in Section~\ref{sec:policy-multi}.}

\item
We conduct numerical experiments in three applications with 2, 3, and 5 modalities. {OPT-ST achieves the lowest inference error among all baselines, with reductions of up to 80.3\%, while requiring less offline computation time than RL. In the five-modality case, RR-ST and RR-Thres achieve $5.2\times$ and $89.8\times$ speedups in offline computation over OPT-ST, respectively, at the cost of increasing the inference error by 23.3\% and 38.6\%, respectively.
We further compare OPT-ST with MGF-Corr, an approximation-based algorithm that uses a weighted sum of scalar AoI functions~\cite{shisher2025aoi}. The results show that, under high correlation, OPT-ST achieves 87.9\% lower inference error than MGF-Corr.}
\end{itemize}


\section{Related Work} 

\begin{table*}[t]
\centering
\caption{Qualitative comparison with the state-of-the-art.}
\label{tab:lit_compare}
\renewcommand{\arraystretch}{1.15}
\setlength{\tabcolsep}{4.5pt}

\begin{tabular}{|L{3.6cm}|L{2.3cm}|L{3.6cm}|L{3.6cm}|L{1.6cm}|L{1.4cm}|}
\hline
\textbf{Goal} 
& \textbf{Source(s)} 
& \textbf{AoI Objective Function} 
& \textbf{Policy} 
& \textbf{Optimality}
& \textbf{Work} \\
\hline

\multirow{2}{3.2cm}{Minimize AoI of a single source}
& \multirow{2}{*}{Single source}
& Linear AoI function 
& Threshold policy
& Optimal
& \cite{8000687} \\
\cline{3-6}

&  
& Non-linear AoI function 
& Threshold policy
& Optimal
& \cite{8764465} \\
\hline

Minimize inference error of a single source
& Single source
& General AoI function 
& Index-based threshold policy
& Optimal
& \cite{shisher2022does,shisher2023learning, ari2026goal, 10559951} \\
\hline

Minimize estimation error of two correlated sources
& Two correlated Gaussian processes 
& Minimum of two non-decreasing AoI functions
& Time-shift scheduling
& Optimal
& \cite{hribar2018using} \\
\hline

Joint design for remote monitoring over random delays
& Two correlated Wiener processes
& Non-decreasing, non-separable function of two AoIs
& Water-Filling (WF) sampler \newline Max Age First scheduler \newline Weighted-sum fusion estimator
& Optimal
& \cite{li2025optimal} \\
\hline

Minimize estimation error of multiple correlated sources
& Multiple correlated Wiener processes
& Weighted sum of AoI functions across sources
& Max-Weight-like policy
& Constant-factor optimal
& \cite{ramakanth2024monitoring} \\
\hline

Minimize inference error of multiple correlated sources over multiple channels
& Multiple correlated sources
& Weighted sum of general AoI-vector functions
& Correlated Net Gain Policy
& Suboptimal
& \cite{shisher2025aoi} \\
\hline

\textbf{Minimize multimodal inference error}
& \textbf{Two correlated sources}
& \textbf{General function of two AoIs}
& \textbf{Index-based threshold policy}
& \textbf{Optimal}
& \textbf{This work}, \cite{11206222} \\
\hline

\textbf{Minimize multimodal inference error}
& \textbf{Multiple correlated sources}
& \textbf{General AoI-vector function}
& \textbf{Switching-type policy}
& \textbf{Optimal}
& \textbf{This work} \\
\hline

\end{tabular}
\end{table*}

\textbf{AoI penalty functions.} \emph{Age of Information (AoI)} is a widely adopted metric for quantifying information freshness (see a comprehensive survey~\cite{yates2021age}). 
Since its inception, the relationship between information \emph{freshness} and its \emph{value} to the application has been evolving~\cite{8764465, zhong2018two, Maatouk2020AoII, sun2018information, chen2022uncertainty, zheng2020urgency, kosta2017age, pappas2021goal, 10356278}.
For example, in~\cite{8764465}, Sun and Cyr suggested several non-decreasing, \emph{non-linear} AoI functions to capture the value of fresh data, since the penalty of stale information is not necessarily linear in time.
More recent research has explored the impact of freshness in remote ML inference systems~\cite{shisher2022does, 10620843, 10559951, shisher2023learning, 9484640, ari2026goal}.
In the seminal work~\cite{9484640, shisher2022does}, Shisher~\emph{et al.} demonstrated that inference error can be expressed as a general AoI penalty function, which may not be monotonic.
Based on this observation, the authors studied a transmission scheduling problem, aiming to minimize a general AoI penalty function.
In~\cite{shisher2023learning}, the joint optimization of feature length selection and transmission scheduling was further studied.
In~\cite{ari2026goal}, Ari~\emph{et al.} studied the minimization of a general AoI penalty function under a two-way delay (transmission and acknowledgment (ACK) delay).
These prior works assume that each task corresponds to a single source, which can be viewed as a \emph{single}-modality setting.
In contrast, this paper studies the case where each task may involve multiple modalities by considering a general function of the AoI vector.
\textbf{AoI optimization for multiple sources.}
This paper is also related to the rich literature of AoI optimization in multi-source systems~\cite{kadota2018scheduling, 8486307, talak2018optimizing, hsu2017scheduling, tripathi2024whittle, tang2020minimizing, li2019kronos, li2025eywa, 9484640, shisher2022does, shisher2023learning, he2018minimizing, yin2020application, tong2022age, tripathi2022optimizing, 10628041, hribar2018using, ramakanth2024monitoring, shisher2025aoi, li2025optimal}.
Various source scheduling policies have been designed to minimize AoI for multiple \emph{independent} sources, such as the threshold policy~\cite{tang2020minimizing}, the Max-weight policy~\cite{8486307}, the cyclic policy~\cite{li2025eywa}, and the Whittle Index scheduling~\cite{kadota2018scheduling, tripathi2024whittle, hsu2017scheduling}.
A common limitation of these approaches is that they assume a one-to-one relationship between sources and the targets of interest~\cite{yin2020application}.

In practice, sources may contain side information about other sources, and the targets of interest may depend jointly on multiple sources.
To address this, several works study AoI optimization for correlated sources~\cite{he2018minimizing, yin2020application, tong2022age, tripathi2022optimizing, 10628041, hribar2018using, ramakanth2024monitoring, shisher2025aoi, li2025optimal}.
The settings most closely related to this paper are remote estimation and remote inference with correlated sources, where the target is estimated or inferred from multiple correlated observations~\cite{hribar2018using, ramakanth2024monitoring, shisher2025aoi, li2025optimal}. In~\cite{hribar2018using}, Hribar~\emph{et al.} studied two correlated Gaussian processes and derived the optimal time shift between periodic updates by modeling the estimation error as the minimum of two non-decreasing AoI functions. In~\cite{ramakanth2024monitoring}, Ramakanth~\emph{et al.} considered multiple correlated discrete-time Wiener processes, bounded the estimation error by weighted sums of sensor AoIs, and proposed a constant-factor optimal Max-Weight-like policy. In~\cite{shisher2025aoi}, Shisher~\emph{et al.} extended this line of work to remote inference with arbitrary error functions and source/target processes, derived an information-theoretic lower bound, and proposed MGF and online policies based on an AoI-based approximation. In~\cite{li2025optimal}, Li and Uysal studied the joint design of sampling, scheduling, and estimation for two correlated continuous-time Wiener processes under random delay.

{The work most closely related to this paper is~\cite{shisher2025aoi}.
Our work differs from~\cite{shisher2025aoi} in the following aspects.
First, the near-optimality guarantee in~\cite{shisher2025aoi} relies on (i) an \emph{optimal} predictor that minimizes the expected inference loss given the available observations and (ii) a specific Markov-chain condition on the source and target processes. Our problem reformulation and algorithm design require neither an optimal predictor nor the Markov property.
Second, in~\cite{shisher2025aoi}, an approximation-based algorithm is developed for a multi-channel system with deterministic unit transmission times. In this paper, we show how to find an optimal policy for a single-channel system with deterministic, heterogeneous transmission and preprocessing times.
Finally, our three-modality simulation in Section~\ref{sec:numerical} shows that our algorithms better handle the complex AoI-vector function in this setting when the correlation is high.}

A preliminary version of this work focusing on the two-modality case appeared in~\cite{11206222}.
We summarize the qualitative comparison between our work and the literature in Table~\ref{tab:lit_compare}.
\section{System Model and Problem Formulation}

\begin{figure}[t!]
    \centering
    \includegraphics[width=0.85\linewidth]{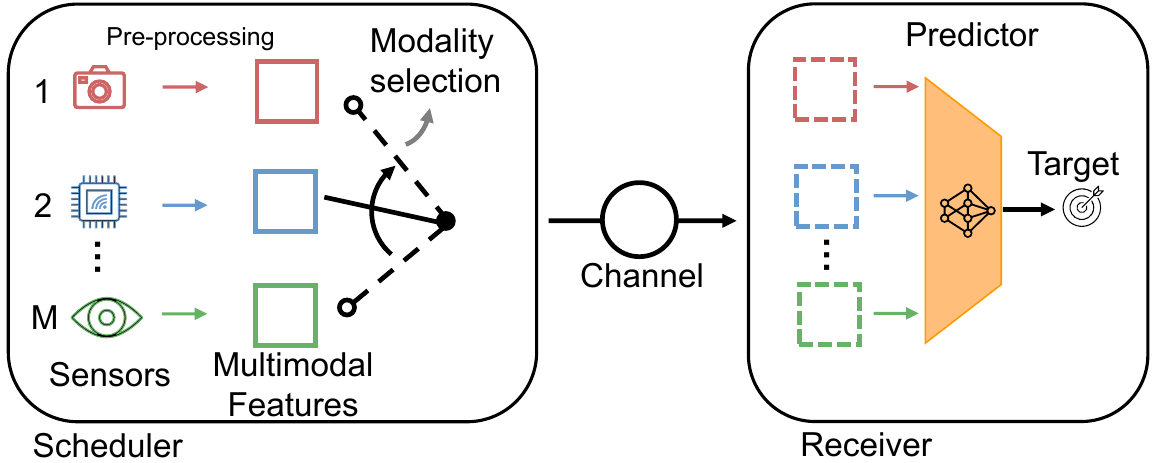}
    \caption{System model.}
    \label{fig:system model}
\end{figure}

In this section, we model the multimodal remote inference system and formulate a modality scheduling problem. We cast the problem as an SMDP and prove the existence of an optimal stationary policy.

As depicted in Fig.~\ref{fig:system model}, we consider a multimodal remote inference system consisting of $M$ sensors, each generating samples of a distinct modality, and a remote inference node that infers the target based on the received features.
Let $\mathcal{M} \coloneq \{1,2,\dots, M\}$ denote the set of modalities.
Time is slotted and indexed by $t = 0,1, \dots$.
{At each time $t$, the sensor for each modality $m$ generates a measurement $Z_{m,t}$.
A raw measurement is transformed into a feature through preprocessing operations such as feature extraction or compression.
Let $T_m^{\mathrm{pre}}$ denote the fixed, modality-dependent preprocessing time for modality $m$.
Thus, the transmitter receives the feature $X_{m,t-T_m^{\mathrm{pre}}}$ from each modality $m$ at time $t$; let $\mathcal{X}_m$ denote modality $m$'s feature space and $\mathcal{X} \coloneq \mathcal{X}_1 \times \cdots \times \mathcal{X}_M$ denote the joint feature space.
When $T_m^{\mathrm{pre}}=0$ and $X_{m,t}=Z_{m,t}$, this means that the transmitter sends raw measurements without preprocessing.}
On the receiver side, a predictor (e.g., a neural network) uses the \emph{freshest received} features from all modalities to infer the current target value $Y_t$ from a given target set $\mathcal{Y}$.

Due to limited network resources, only one modality can be transmitted at a time, and each transmission may span multiple time slots.
Suppose the $i$-th transmission starts at time $S_i$ and completes at time $D_i$.
At time $S_i$, let $d_i \in \mathcal{M}$ denote the modality selected for transmission.
Then, the transmitter begins transmitting the freshest feature from modality $d_i$, namely \(X_{d_i,S_i-T_{d_i}^{\mathrm{pre}}}\).
Let $T^{\mathrm{tx}}_m$ denote the \emph{fixed} transmission time for modality $m$, which is a positive finite integer.
Because modalities may have different feature sizes, their transmission times \(T^{\mathrm{tx}}_1,\dots, T^{\mathrm{tx}}_M\) may differ.
We assume that the channel is reliable and transmissions are non-preemptive.
That is, the receiver always successfully receives the feature from modality $d_i$ at time $D_i$, and the transmission duration is $T^{\mathrm{tx}}_{d_i}$ (i.e., $D_i - S_i = T^{\mathrm{tx}}_{d_i}$).
We also assume a work-conserving system: when the current transmission completes, the subsequent transmission begins immediately, i.e., $S_{i+1} = D_i$ for every $i$.


\subsection{Age of Information}

We use Age of Information (AoI) to quantify the freshness of features at the receiver, defined as the time elapsed since the freshest received feature was generated~\cite{8000687}.
Let $\Delta_m(t) \in \mathbb{N}_+$ denote the AoI of modality $m$ at the receiver at time $t$, where $\mathbb{N}_+$ is the set of positive integers.
{For each modality $m$, whenever the receiver receives a feature of modality $m$ at $t=D_i$, where $d_i=m$, its AoI resets to the age of the received feature, i.e., $T_m^{\mathrm{pre}} + T_m^{\mathrm{tx}}$; otherwise, it increases by one.
That is, the AoI of each modality $m$ evolves as
\begin{equation}
    \Delta_m(t) = \begin{cases}
        T_m^{\mathrm{pre}} + T_m^{\mathrm{tx}} & \!\text{if} \ t=D_i \ \text{and} \  d_i=m, \\
         \Delta_m(t-1) + 1 & \!\text{otherwise}.
    \end{cases}
\end{equation}
}
The AoI vector $\boldsymbol{\Delta}(t)$ of all modalities is defined as
\[
\boldsymbol{\Delta}(t) \coloneq \left (\Delta_1(t), \Delta_2(t), \dots, \Delta_M(t) \right) \in \mathbb{N}_+^M.
\]

For technical reasons, our analysis and algorithm design require a finite AoI vector space. We therefore approximate the true AoI evolution by considering a truncated AoI vector space.
Specifically, define the operator $\wedge$ by $x \wedge y \coloneq \min\{x, y\}$.
{
For each modality $m$, let $K_m$ be a sufficiently large integer. The truncated AoI evolution is given by
\begin{equation} \label{eq:AoI-evolution}
    \Delta_m(t) = \begin{cases}
        T_m^{\mathrm{pre}} + T_m^{\mathrm{tx}} & \!\text{if} \ t=D_i \ \text{and} \  d_i=m, \\
         (\Delta_m(t-1) + 1) \wedge K_m & \!\text{otherwise}.
    \end{cases}
\end{equation}
Define the truncation vector as $\boldsymbol{K}\coloneqq(K_1,\ldots,K_M)$ and the truncated AoI space as $\mathcal{K}\coloneqq\prod_{m\in\mathcal{M}}\{1,\ldots,K_m\}$.
Intuitively, when $K_m$ is large, the inaccuracy induced by state truncation vanishes.
Throughout the remainder of the paper, unless otherwise stated, all AoI expressions are interpreted with componentwise truncation at $\boldsymbol{K}$.
}


\subsection{Predictor and AoI-based Inference Error}
Denote the freshest received feature as $\mathbf{X}_{t - \boldsymbol{\Delta}(t)}$, defined as
\[
\mathbf{X}_{t - \boldsymbol{\Delta}(t)} \coloneq (X_{1, t-\Delta_1(t)}, X_{2, t - \Delta_2(t)}, \dots, X_{M, t - \Delta_M(t)}).
\]
In order to predict the target $Y_t\in\mathcal Y$, the predictor $\phi: \mathcal{X} \times \mathcal{K} \to \mathcal{A}$  outputs an action $A_t$ from a given action set $\mathcal A$; the action is determined based on the freshest received features $\mathbf{X}_{t - \boldsymbol{\Delta}(t)}\in \mathcal{X}$ and the associated AoI vector $\boldsymbol{\Delta}(t)\in \mathcal{K}$.
The prediction is evaluated using a loss function 
$\ell(y, a)$, which quantifies the inference error incurred if the predictor selects action $a\in\mathcal A$ while the true target value is $y\in\mathcal Y$.
For example, the action can be a probability distribution $Q_Y$ in the space $\mathcal Y$, with the associated logarithmic loss function $\ell_{\mathrm{log}}(y, Q_Y) \coloneq -\log Q_Y(y)$.
Alternatively, the action can be an estimate $\hat{y} \in \mathcal{Y}$  of the true target value  $y\in\mathcal Y$, with the associated quadratic loss function $\ell_{2}(y, \hat{y}) \coloneq (y - \hat{y})^2$.

{
During online inference, the predictor $\phi$ predicts the target value at each time slot. The expected inference error at time $t$ is defined as
\begin{equation}\label{eq:expected_inference_error}
\mathcal{E}(t)
\coloneqq
\mathbb{E}_{
(Y,\mathbf{X},\boldsymbol{\delta})
\sim
\mathbb{P}_{
Y_t,\mathbf{X}_{t-\boldsymbol{\Delta}(t)},\boldsymbol{\Delta}(t)
}}
\left[
\ell\left(Y,\phi(\mathbf{X},\boldsymbol{\delta})\right)
\right],
\end{equation}
where $\mathbb{P}_{Y_t,\mathbf{X}_{t-\boldsymbol{\Delta}(t)},\boldsymbol{\Delta}(t)}$ denotes the joint distribution of the target, the features, and the AoI vector at time $t$.
In general, this distribution may depend on $Y_t$, $\mathbf{X}_t$, and $\boldsymbol{\Delta}(t)$, making the expected inference error difficult to optimize directly. It was first shown in~\cite{9484640,shisher2022does} that, under certain assumptions, the expected inference error can be expressed as the expectation of an AoI-vector function. Specifically, consider the following two assumptions:
\begin{assumption}\label{ass:stationary}
The process $\{(Y_t,\mathbf{X}_t),\,t\in \mathbb{Z}\}$ is stationary.
\end{assumption}
\begin{assumption}\label{ass:independence}
The process $\{(Y_t,\mathbf{X}_t),\,t\in \mathbb{Z}\}$ is independent of the process $\{\boldsymbol{\Delta}(t),\,t\in \mathbb{Z}\}$.
\end{assumption}

Assumption~\ref{ass:stationary} means that the relationship between the features and the target does not change over time.
Assumption~\ref{ass:independence} holds under the signal-agnostic scheduling considered in this paper, where scheduling decisions are based on the AoI process rather than the realized feature or target values.
This setting is consistent with prior studies on AoI-based remote inference~\cite{shisher2022does,10559951,shisher2023learning,ari2026goal,shisher2025aoi}.
For every AoI vector $\boldsymbol{\delta}\in\mathcal{K}$, define the function
\begin{equation}\label{eq:inference error}
L(\boldsymbol{\delta})
\coloneqq
\mathbb{E}_{
(Y,\mathbf{X})
\sim
\mathbb{P}_{Y_t,\mathbf{X}_{t-\boldsymbol{\delta}}}
}
\left[
\ell\left(Y,\phi(\mathbf{X},\boldsymbol{\delta})\right)
\right],
\end{equation}
where $\mathbb{P}_{Y_t,\mathbf{X}_{t-\boldsymbol{\delta}}}$ denotes the joint stationary distribution of the target and the features under AoI vector $\boldsymbol{\delta}$.
\begin{lemma}[\!\cite{9484640,shisher2022does}]\label{lem:L}
Under Assumptions~\ref{ass:stationary} and~\ref{ass:independence}, for a given predictor $\phi$, the expected inference error at time $t$ in Eq.~\eqref{eq:expected_inference_error} is given by
\begin{equation}\label{eq:error_as_L}
\mathcal{E}(t)
=
\mathbb{E}_{\boldsymbol{\Delta}(t)}[L\left(\boldsymbol{\Delta}(t)\right)].
\end{equation}
\end{lemma}

\begin{proof}
See~\cite[Appendix VIII-D]{shisher2022does} for the proof.
\end{proof}

Lemma~\ref{lem:L} suggests that, instead of directly optimizing the expected inference error, we can optimize the expectation of an AoI-vector function.
Function $L$ can be quite general, as it depends jointly on the predictor $\phi$, the loss function $\ell$, and the statistics of the feature and target processes.}
We only assume that $L$ is uniformly bounded, as stated below.
\begin{assumption} \label{ass:bounded}
There exists a real number $\alpha > 0$ such that $|L(\boldsymbol{\delta})| \le \alpha$ for all $\boldsymbol{\delta}\in\mathcal{K}$.
\end{assumption}
This assumption is practical, as preprocessing techniques in ML applications typically bound the inference error.
We assume that the scheduler knows the function $L$.
{For a data-driven ML inference model, training and evaluation typically involve three stages: training, validation, and testing.
The function $L$ can be obtained by averaging the prediction losses over validation samples constructed with each AoI vector.
We will discuss how to obtain $L$ in detail in Section~\ref{sec:numerical}.}

\subsection{Problem Formulation}

We formulate the scheduling problem as a semi-Markov decision process (SMDP), because decisions are made only when a transmission is completed, and the interval between consecutive decision epochs depends on the decisions.

\textbf{State:} At each decision epoch $S_i$, the system state $s_i$ is the AoI vector $\boldsymbol{\Delta}(S_i)$ and the state set is $\mathcal{S} \coloneq \mathcal{K}$.

\textbf{Decision:} The decision is the modality choice $d_i$ and the decision set is $\mathcal{D} \coloneq \mathcal{M}$.

\textbf{Transition time:} If modality \(d_i\) is selected at time \(S_i\), the next decision epoch occurs upon completion of its transmission. Hence, the transition time is $\tau_i \coloneq S_{i+1} - S_i = T^{\mathrm{tx}}_{d_i}$.

\textbf{State transition:} According to the AoI evolution in Eq.~\eqref{eq:AoI-evolution}, the AoI of modality \(m\) in the next state satisfies
\begin{equation} \label{eq:state-transition}
    \Delta_m(S_{i+1}) = \begin{cases}
         T_m^{\mathrm{tx}}+T_m^{\mathrm{pre}} & \!\text{if} \ m = d_i, \\
         \Delta_m(S_{i}) + T^{\mathrm{tx}}_{d_i} & \!\text{if} \ m \neq d_i.
    \end{cases}
\end{equation}
{Because the transmission time is deterministic, we have
\begin{equation}
\mathbb{P}\!\left(
\boldsymbol{\Delta}(S_{i+1}), \tau_i
\mid
\boldsymbol{\Delta}(S_i), d_i
\right) = 1,
\end{equation}
which means that, conditioned on the current AoI state
\(\boldsymbol{\Delta}(S_i)\) and action \(d_i\), the next AoI state is
\(\boldsymbol{\Delta}(S_{i+1})\) and the transmission time is \(\tau_i\) with
probability one.}

\textbf{Cost function:} At each time slot $t$, the instantaneous cost is the expected inference error $L(\boldsymbol{\Delta}(t))$. Therefore, the cumulative cost incurred during the transmission interval $[S_i, S_{i+1})$ is $c\bigl(\boldsymbol{\Delta}(S_i), d_i, \tau_i\bigr) \coloneq \sum_{t=0}^{\tau_i-1} L\bigl(\boldsymbol{\Delta}(S_i+t)\bigr)$.

\textbf{Scheduling policy.} We consider \emph{signal-agnostic} scheduling: at each time, the scheduler observes the AoI vector.
Let $\pi_i$ denote a decision rule that assigns a probability distribution over modalities at decision epoch $S_i$ based on the history of observed AoI vectors and past scheduling decisions.
A scheduling policy is then a sequence of decision rules, i.e., $\pi \coloneq (\pi_0,\pi_1,\dots)$. Let $\Pi$ denote the set of all such history-dependent randomized policies.




\textbf{Objective.} For a policy $\pi \in \Pi$ and an initial AoI state $\boldsymbol{\Delta}(0) = \boldsymbol{\delta}$, the time average expected inference error over an infinite horizon is defined as
\begin{equation}\label{eq:criterion}
    \bar{L}_{\pi}(\boldsymbol{\delta}) = \limsup_{T\to\infty} \frac{1}{T} \mathbb{E}_{\pi}\left[\sum_{t=0}^{T-1} L\left( \boldsymbol{\Delta}(t) \right ) \Big| \boldsymbol{\Delta}(0) = \boldsymbol{\delta} \right].
\end{equation}
{
We aim to minimize the time-average expected inference error:
\begin{equation}\label{eq:original-problem}
    \bar{L}_{\mathrm{opt}}(\boldsymbol{\delta}) \coloneqq \inf_{\pi\in\Pi}\bar{L}_{\pi}(\boldsymbol{\delta}),
    \qquad \boldsymbol{\delta}\in\mathcal{K}.
\end{equation}
}

\subsection{Existence of an Optimal Stationary Policy}
We are particularly interested in stationary policies because of their simplicity.
A policy is called \emph{stationary} if it uses the same deterministic decision rule at every decision epoch.\footnote{There is no general agreement on whether ``stationary'' implies deterministic~\cite[Chapter 1.2]{feinberg2012handbook}. For simplicity, we assume that stationary policies are deterministic throughout the paper.}
Let $\Pi^{\mathrm S} \subseteq \Pi$ denote the set of stationary policies.
In general, an optimal stationary policy may not exist for average-reward SMDPs~\cite[Examples 8.10.1–8.10.2]{puterman2014markov}. Lemma~\ref{lem:existence} shows that an optimal stationary policy does exist in our setting.

\begin{lemma} \label{lem:existence}
    {For problem~\eqref{eq:original-problem}, when the state set, decision set, and transmission times are finite and the inference error function $L$ is uniformly bounded,} there exists a stationary policy $\pi^* \in \Pi^{\mathrm S}$ such that $\bar{L}_{\pi^*}(\boldsymbol{\delta}) = \bar{L}_{\mathrm{opt}}(\boldsymbol{\delta})$ for every $\boldsymbol{\delta} \in \mathcal{K}$.
\end{lemma}
\begin{proofsketch}
We prove this lemma by applying Theorem~3 in~\cite{yushkevich1982semi}, which provides a sufficient condition for the existence of an optimal stationary policy. In particular, our model satisfies the condition because the transmission times are finite and discrete, and the inference error function is uniformly bounded.
We provide the complete proof in Appendix~\ref{app:existence}.
\end{proofsketch}

Lemma~\ref{lem:existence} allows us to focus on finding an optimal stationary policy. In the next section, we show that this is equivalent to finding an optimal stationary policy for a reformulated problem that is more computationally efficient.

\section{Problem Reformulation}\label{sec:optimal-policy}


{We first identify a simple case that has an optimal fixed-modality policy. For the remaining case, we formulate an SMDP on a reduced state set and prove its equivalence with the original problem. We then characterize its chain structure and present the Bellman optimality equations used to derive scheduling policies.}


\begin{figure}[t]
\centering
\includegraphics[width=0.60\linewidth]{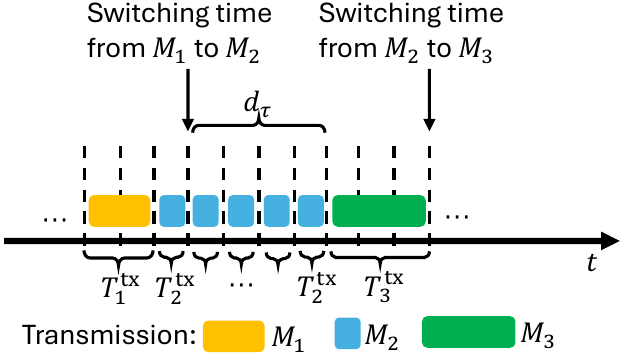}
\caption{Illustration of switching times. At the switching time from $M_1$ to $M_2$, the scheduler selects the decision pair $(d_\tau, d_s)$. In this example, we have $d_\tau = 4$ and $d_s = M_3$.}
\label{fig:switching_time}
\end{figure}

\subsection{Modality Switching Times} \label{sec:switching-case}
Recall that the scheduler selects a modality at the start of each transmission.
We call the delivery time \(D_i\) a \emph{switching time} if transmissions \(i-1\) and \(i\) use different modalities; see Fig.~\ref{fig:switching_time} for an example. 
{
At each switching time, the AoI values of two modalities are known, reducing the state space.
For a switch from modality \(m\) to modality \(n\), where \(m \neq n\), the feature of modality \(n\) has just been delivered, while that of modality \(m\) was delivered \(T^{\mathrm{tx}}_n\) time slots earlier.
Thus, modality \(n\) has AoI \(T^{\mathrm{pre}}_n + T^{\mathrm{tx}}_n\), and modality \(m\) has AoI \(T^{\mathrm{pre}}_m + T^{\mathrm{tx}}_m + T^{\mathrm{tx}}_n\).

The concept of switching times raises a question: should the scheduler keep switching modalities, or eventually settle on one?
To answer this question, we introduce the following condition:
\begin{condition}\label{con:reduction}
There exist an optimal stationary policy $\pi^*$ for the original
problem~\eqref{eq:original-problem} and an initial AoI state $\boldsymbol{\delta}_0 \in \mathcal{S}$
such that, starting from $\boldsymbol{\delta}_0$, the scheduler switches
modalities infinitely many times under $\pi^*$.
\end{condition}

The following lemma shows that the original problem~\eqref{eq:original-problem} can be solved directly when Condition~\ref{con:reduction} does not hold.
For brevity, we use $\boldsymbol{\delta}_{-m}$ to denote the AoI vector $\boldsymbol{\delta}$ without its $m$-th entry, and $(x;\boldsymbol{\delta}_{-m})$ to denote $\boldsymbol{\delta}$ with that entry replaced by $x$.
Similarly, $\boldsymbol{\delta}_{-(m,n)}$ denotes $\boldsymbol{\delta}$ without its $m$-th and $n$-th entries, and $(x,y;\boldsymbol{\delta}_{-(m,n)})$ denotes $\boldsymbol{\delta}$ with those entries replaced by $x$ and $y$, respectively.

\begin{lemma} \label{lem:condition}
If Condition~\ref{con:reduction} does not hold, then there exists an
optimal stationary policy $\pi^*$ for the original problem~\eqref{eq:original-problem} that always selects a fixed modality $m^*$, which is given by
\begin{equation}
\label{eq:optm-special}
    m^* \in \arg\min_{m\in\mathcal{M}}
    \frac{1}{T^{\mathrm{tx}}_m}
    \sum_{t=0}^{T^{\mathrm{tx}}_m-1}
    L\left(
        T^{\mathrm{pre}}_m + T^{\mathrm{tx}}_m+t;
        \boldsymbol{K}_{-m}
    \right).
\end{equation}
Accordingly, the optimal objective value of problem~\eqref{eq:original-problem} is
\begin{equation}
\label{eq:optimal-value-special}
    \bar{L}_{\mathrm{opt}}(\boldsymbol{\delta})
    =
    \frac{1}{T^{\mathrm{tx}}_{m^*}}
    \sum_{t=0}^{T^{\mathrm{tx}}_{m^*}-1}
    L\left(
        T^{\mathrm{pre}}_{m^*} + T^{\mathrm{tx}}_{m^*}+t;
        \boldsymbol{K}_{-m^*}
    \right),
\end{equation}
where $\boldsymbol{\delta}$ is any initial AoI vector. $\boldsymbol{K}$ is the truncation vector.
\end{lemma}
\begin{proofsketch}
If Condition~\ref{con:reduction} does not hold, an optimal stationary policy switches modalities only finitely many times from any initial AoI state.
Hence, after some finite time, it always selects the same modality.
The AoIs of all other modalities then reach their truncation limits.
Since the finite transient does not affect the long-run average cost, it suffices to select the fixed modality with the minimum average cost.
\end{proofsketch}

Lemma~\ref{lem:condition} shows that, when Condition~\ref{con:reduction} does not hold, the original problem~\eqref{eq:original-problem} can be solved by comparing the $M$ fixed-modality policies.
We now consider the case where Condition~\ref{con:reduction} holds.
To solve the problem in this case, we next formulate an SMDP using switching times as decision epochs.
}

\subsection{SMDP Problem Reformulation} \label{sec:def-reform}
The reformulated problem has five components: the state set $\mathcal{S}_\mathrm{re}$, the decision set $\mathcal{D}_\mathrm{re}$, the transition time $\tau_\mathrm{re}$, the transition probability distribution $\mathbb{P}_\mathrm{re}$, and the cost function $c_\mathrm{re}(s_\mathrm{re}, d_\mathrm{re})$ depending on state $s_\mathrm{re} \in \mathcal{S}_\mathrm{re}$ and decision $d_\mathrm{re} \in \mathcal{D}_\mathrm{re}$.

\textbf{State:} {For any two distinct sources \(m,n\in\mathcal{M}\) ($m \neq n$), each possible AoI state at
a switching time from source \(m\) to source \(n\) can be represented as
\begin{equation}\label{eq:def-switch}
    s_{m\to n}^{\boldsymbol{\delta}}
    \coloneq
    (T^{\mathrm{pre}}_m + T^{\mathrm{tx}}_m + T^{\mathrm{tx}}_n, T^{\mathrm{pre}}_n + T^{\mathrm{tx}}_n; \boldsymbol{\delta}_{-(m,n)}),
\end{equation}
where \(\boldsymbol{\delta}\) is an arbitrary AoI vector. Given $m, n$, all possible AoI states associated with this switching pair
form the set
\begin{equation}\label{eq:switching-state-set}
    \mathcal{S}_{m\to n}
    \coloneq
    \left\{
    s_{m\to n}^{\boldsymbol{\delta}}
    \mid
    \boldsymbol{\delta}\in\mathcal{K}
    \right\}.
\end{equation}
Hence, the reformulated state set is the union of sets over all ordered pairs of
distinct sources:
\begin{equation} \label{eq:reformulated-state}
    \mathcal{S}_\mathrm{re}
    \coloneq
    \bigcup_{\substack{m,n\in\mathcal{M}\\ m\neq n}}
    \mathcal{S}_{m\to n}.
\end{equation}
}


\textbf{Decision:} At each switching state, the scheduler makes two decisions: (\romannumeral 1) the finite number of consecutive transmissions of the current modality, and (\romannumeral 2) the next modality to switch to.
Let \(d_{\mathrm{re}} \coloneq (d_\tau, d_s)\) denote the reformulated decision, where \(d_\tau\) specifies decision~(\romannumeral 1) and \(d_s\) specifies decision~(\romannumeral 2).
Given any switching state $s_{m \to n}^{\boldsymbol{\delta}}$, the reformulated decision set is $\mathcal{D}_{\mathrm{re}}(s_{m\to n}^{\boldsymbol{\delta}})\coloneq \mathbb{N}\times (\mathcal{M}\setminus\{n\})
$.
Note that the current modality $n$ is excluded because $d_s$ must select a different modality.

{
\textbf{Transition time:}
With reformulated state $s_\mathrm{re} = s_{m\to n}^{\boldsymbol{\delta}}$ and reformulated decision $d_{\mathrm{re}}$, the transition time is given by
\begin{equation} \label{eq:transition-time}
    \tau_{\mathrm{re}}(s_\mathrm{re}, d_{\mathrm{re}}) = d_\tau T^{\mathrm{tx}}_n + T^{\mathrm{tx}}_{d_s},
\end{equation}
because there are $d_\tau$ transmissions of modality $n$, followed by one transmission of modality $d_s$, as illustrated in Fig.~\ref{fig:switching_time}.

\textbf{State transition:}
Furthermore, let $s'_{\mathrm{re}}$ denote the next reformulated state after $s_\mathrm{re}$, which has three parts:
First, the AoI of modality $n$ is $T_n^{\mathrm{pre}}+T_n^{\mathrm{tx}}+T_{d_s}^{\mathrm{tx}}$, 
since it increases from $T_n^{\mathrm{pre}}+T_n^{\mathrm{tx}}$ by $T_{d_s}^{\mathrm{tx}}$.
Second, the AoI of modality $d_s$ is $T_{d_s}^{\mathrm{pre}}+T_{d_s}^{\mathrm{tx}}$, 
since its feature has just been delivered. 
Third, the AoIs of all other modalities increase by the total transition time,
since they are not updated.
Formally, let $\delta'_k$ denote the AoI of modality $k$ in the reformulated state $s'_{\mathrm{re}}$. Then, we have
\begin{equation}
\label{eq:next-state}
\delta_k' =
\begin{cases}
T_n^{\mathrm{pre}} + T_n^{\mathrm{tx}} + T_{d_s}^{\mathrm{tx}},
    & k=n, \\[2pt]
T_{d_s}^{\mathrm{pre}} + T_{d_s}^{\mathrm{tx}},
    & k=d_s, \\[2pt]
\delta_k + \tau_{\mathrm{re}}(s_{\mathrm{re}},d_{\mathrm{re}}),
    & k\notin\{n,d_s\},
\end{cases}
\end{equation}
where $\delta_k$ is modality $k$'s AoI in the current reformulated state $s_{\mathrm{re}}=s^{\boldsymbol{\delta}}_{m\to n}$.
The transition probability distribution $\mathbb{P}_{\mathrm{re}}$ is
\begin{equation} \label{eq:transition-prob}
    \mathbb{P}_{\mathrm{re}}\!\left(j, \tau \mid s_\mathrm{re}, d_\mathrm{re} \right) =
    \begin{cases}
        1 & \text{if } j = s'_\mathrm{re} \ \text{and} \ \tau = \tau_{\mathrm{re}}(s_\mathrm{re}, d_\mathrm{re}), \\
        0 & \text{otherwise.}
    \end{cases}
\end{equation}

\textbf{Cost function:} The transition cost is the cumulative expected inference error during $d_\tau$ consecutive transmissions of modality $n$ and the subsequent transmission of modality $d_s$.
Writing the current AoI state as $s_{\mathrm{re}}=\boldsymbol{\delta}$ and the decision as $d_{\mathrm{re}}=(d_\tau,d_s)$, we have
\begin{equation}\label{eq:reformulated-cost}
\begin{aligned}
&c_{\mathrm{re}}(s_{\mathrm{re}},d_{\mathrm{re}})
\\
&=\sum_{k=0}^{d_\tau-1}\sum_{t=0}^{T_n^{\mathrm{tx}}-1}
L\!\bigl(\delta_n+t;\boldsymbol{\delta}_{-n}+(kT_n^{\mathrm{tx}}+t)\mathbf{1}_{M-1}\bigr)
\\
&\quad+\sum_{t=0}^{T_{d_s}^{\mathrm{tx}}-1}
L\!\bigl(\delta_n+t;\boldsymbol{\delta}_{-n}+(d_\tau T_n^{\mathrm{tx}}+t)\mathbf{1}_{M-1}\bigr).
\end{aligned}
\end{equation}
Here, $\delta_n=T_n^{\mathrm{pre}}+T_n^{\mathrm{tx}}$ is the AoI of modality $n$ at the current switching time, and $\mathbf{1}_{M-1}$ is the all-ones vector of length $M-1$. All AoI arguments are truncated componentwise at $\boldsymbol{K}$.
The first sum is zero when $d_\tau=0$.
The derivation of Eq.~\eqref{eq:reformulated-cost} is provided in Appendix~\ref{app:transition-cost}.
}

{
\textbf{Objective:} Let $\pi_{\mathrm{re}}: \mathcal{S}_{\mathrm{re}} \to \mathcal{D}_{\mathrm{re}}$ be a stationary reformulated policy that assigns a feasible decision to each state, and let $\Pi_{\mathrm{re}}^{\mathrm{S}}$ denote the class of such policies.
Let $s_{\mathrm{re},i}$ and $d_{\mathrm{re},i}=\pi_{\mathrm{re}}(s_{\mathrm{re},i})$ be the reformulated state and decision at the $i$-th reformulated decision epoch, and let $N(T)$ denote the number of completed intervals between decision epochs by time $T$.
For each $\pi_{\mathrm{re}}\in\Pi_{\mathrm{re}}^{\mathrm S}$
and $s\in\mathcal{S}_{\mathrm{re}}$, define
\begin{equation}\label{eq:reformulated-criterion}
\begin{aligned}
    &\bar{L}_{\pi_{\mathrm{re}}}(s)\\
    &\quad\coloneqq
    \limsup_{T\to\infty}\frac{1}{T}
    \mathbb{E}_{\pi_{\mathrm{re}}}\!\left[
        \sum_{i=0}^{N(T)-1}
        c_{\mathrm{re}}(s_{\mathrm{re},i},d_{\mathrm{re},i})
        \,\middle|\, s_{\mathrm{re},0}=s
    \right].
\end{aligned}
\end{equation}
The reformulated problem is to minimize the time-average expected inference error:
\begin{equation}\label{eq:reformulated-problem}
    \bar{L}_{\mathrm{opt,re}}(s)
    \coloneqq
    \inf_{\pi_{\mathrm{re}}\in\Pi_{\mathrm{re}}^{\mathrm S}}
    \bar{L}_{\pi_{\mathrm{re}}}(s).
\end{equation}
Note that we omit the cost incurred from the last decision epoch up to time $T$; this cost is bounded under any fixed stationary policy, so its contribution to the time-average cost vanishes as $T\to\infty$.
}

{
\subsection{Equivalence with the Original Problem} \label{sec:equivalence}
The following theorem establishes the equivalence between the original problem~\eqref{eq:original-problem} and the reformulated problem~\eqref{eq:reformulated-problem} under Condition~\ref{con:reduction}.

\begin{theorem}\label{thm:equivalence}
Under Condition~\ref{con:reduction}, the reformulated problem~\eqref{eq:reformulated-problem}
has an optimal stationary policy
$\pi_{\mathrm{re}}^* \in \Pi_{\mathrm{re}}^{\mathrm{S}}$
that achieves a constant optimal objective value, i.e.,
\begin{equation}\label{eq:reformulated-optimal-value}
    \bar{L}_{\pi_{\mathrm{re}}^*}(\boldsymbol{\delta}_{\mathrm{re}})
    = \bar{L}_{\mathrm{opt,re}},
    \quad
    \forall\,\boldsymbol{\delta}_{\mathrm{re}}\in\mathcal{S}_{\mathrm{re}}.
\end{equation}
Furthermore, this constant is also the optimal objective value of the original problem~\eqref{eq:original-problem}, i.e.,
\begin{equation}\label{eq:equivalence}
    \bar{L}_{\mathrm{opt}}(\boldsymbol{\delta})
    = \bar{L}_{\mathrm{opt,re}},
    \quad
    \forall\,\boldsymbol{\delta}\in\mathcal{K}.
\end{equation}
\end{theorem}
\begin{proofsketch}
The proof proceeds in three steps. First, under Condition~\ref{con:reduction}, we construct a stationary reformulated policy with the same average cost as an optimal original policy from an initial state that leads to infinitely many switches.
Second, we use the recurrent cycle reached by this policy to construct a unichain reformulated policy with the same average cost from every switching state. We prove its optimality by showing that a lower-cost reformulated policy would contradict the optimality of the original policy.
Finally, we extend the constructed policy to every original initial state and prove that it remains optimal, establishing Eq.~\eqref{eq:equivalence}.
The complete proof is in Appendix~\ref{app:equivalence}.

\end{proofsketch}
}

According to Eq.~\eqref{eq:equivalence} in Theorem~\ref{thm:equivalence}, the two problems have the same optimal value under Condition~\ref{con:reduction}.
If the initial state lies outside $\mathcal{S}_{\mathrm{re}}$, the scheduler can first transmit two distinct modalities to reach a switching state and then follow $\pi_{\mathrm{re}}^*$.
This finite transient does not affect the average cost. Therefore, it suffices to solve the reformulated problem.



\subsection{Chain Structure of SMDP}\label{sec:chain-structure}

{To solve the reformulated problem, we first examine its chain structure, which determines the form of the Bellman optimality equation.} We recall the definitions of unichain and multichain SMDPs in average-cost Markov decision process theory.

\begin{definition}[\!\!{\cite[Ch. 8.3.1]{puterman2014markov}}]
A finite-state SMDP is called
\begin{enumerate}[label=(\roman*)]
    \item \textbf{unichain}, if the Markov chain corresponding to every stationary policy consists of a single recurrent class plus a possibly empty set of transient states;
    \item \textbf{multichain}, if the Markov chain corresponding to some stationary policy contains two or more recurrent classes.\footnote{For definitions of recurrent and transient classes for finite-state Markov chains, see~\cite[Appendix~A]{puterman2014markov} and~\cite[Chapter~4]{gallager2013stochastic}.}
\end{enumerate}
\end{definition}
The next lemma shows that the chain structure of the reformulated problem is fundamentally different in the two-modality and multi-modality cases.
\begin{lemma}\label{lem:chain_structure}
When \(M=2\), the SMDP of the reformulated problem~\eqref{eq:reformulated-problem} is unichain; when \(M>2\), it can be multichain.
\end{lemma}
\begin{proofsketch}
    For $M=2$, we use the result stated in~\cite[Exercise 4.3]{gallager2013stochastic}: \emph{If a finite-state Markov chain has a state that is accessible from every other state, then the Markov chain has exactly one recurrent class.} We show that such states exist when $M=2$.
    When $M>2$, we construct counterexamples showing that the SMDPs of the reformulated and original problems can be multichain. We provide the complete proof in Appendix~\ref{app:chain}.
\end{proofsketch}

\begin{remark}
Intuitively, when $M=2$, each modality has a unique switching state, and both switching states are accessible from every other state. When $M>2$, each modality may have multiple switching states, and the induced chain may therefore be multichain. Some prior work on AoI optimization for correlated sources applies the unichain Bellman optimality equation after decomposing the problem into single-source subproblems (e.g.,~\cite{tong2022age}). In the multimodal setting, however, the unichain Bellman optimality equation may not apply.
\end{remark}
{
\begin{remark} \label{remark:multichain}
Note that, in the proof of Theorem~\ref{thm:equivalence}, we show that there exists a unichain optimal policy for the reformulated problem~\eqref{eq:reformulated-problem}. This does not contradict the conclusion that the problem can be multichain when $M>2$, because a non-optimal policy may still be multichain. This multichain result is important for the design of our subsequent policy-iteration algorithm: before convergence, an intermediate policy may be non-optimal and multichain.
\end{remark}
}

\subsection{Bellman Optimality Equation}\label{sec:bellman-optimality}

We present the Bellman optimality equations for multichain and unichain SMDPs and explain the difference between them.

For an SMDP specified by $(\mathcal{S}, \mathcal{D}, \tau, \mathbb{P}, c(\cdot))$, the Bellman optimality equation is
\begin{subequations}\label{eq:general-bellman}
\begin{equation}\label{eq:general-g}
g(s)=\min_{d\in\mathcal{D}(s)} \sum_j\sum_\tau \mathbb{P}(j, \tau \mid s,d)\, g(j),
\end{equation}
\begin{equation}\label{eq:general-h}
h(s)=\min_{d\in\mathcal{D}^*(s)} \sum_{j,\tau}\mathbb{P}(j,\tau\mid s,d)\big[c(s,d,\tau)-g(j)\tau+h(j)\big],
\end{equation}
\end{subequations}
for each state $s$.
The set $\mathcal{D}^*(s)$ consists of the decisions attaining the minimum in Eq.~\eqref{eq:general-g}.
{Here, $\mathbb{P}(j,\tau\mid s,d)$ is the probability that the next transition is to state $j$ after $\tau$ time slots, given decision $d$ in state $s$.
The function $g(s)$ represents the optimal average cost starting from state $s$.
The function $h(s)$ is the relative value function, which measures the difference between the optimal expected total cost starting from state $s$ and the accumulated optimal average cost.}
When the SMDP is unichain, the function \(g(s)\) reduces to a constant independent of the state \(s\)~\cite{puterman2014markov}. Let \(g\) denote this constant. Then, Eq.~\eqref{eq:general-g} becomes redundant, because for any decision $d$, we have
\[
    \sum_j \sum_\tau \mathbb{P}(j,\tau \mid s,d)\, g
    = g \cdot \sum_j \sum_\tau \mathbb{P}(j,\tau \mid s,d)
    = g,
\]
and the Bellman optimality equation reduces to
\begin{equation}\label{eq:unichain-bellman}
    h(s)=\min_{d\in\mathcal{D}(s)} \sum_{j,\tau}\mathbb{P}(j,\tau\mid s,d)\big[c(s,d,\tau)-g\tau+h(j)\big],
\end{equation}
for each state $s$. Because $g$ is constant and the Bellman equation is simpler, the unichain SMDP is easier to solve.

An optimal stationary policy is obtained by selecting decisions that attain the minimum in Eq.~\eqref{eq:general-bellman} or Eq.~\eqref{eq:unichain-bellman}.

{For the reformulated problem, the state transition and transition time are deterministic. Hence, Eq.~\eqref{eq:general-bellman} becomes}
\begin{subequations}\label{eq:Bellman-equation}
\begin{equation}\label{eq:multichain-bellman-g}
g(s)=\min_{d\in\mathcal{D}_{\mathrm{re}}(s)} g(s'), \quad s\in\mathcal{S}_{\mathrm{re}},
\end{equation}
\begin{equation}\label{eq:multichain-bellman-h}
h(s)=\min_{d\in\mathcal{D}_{\mathrm{re}}^*(s)}
\big[ c_{\mathrm{re}}(s,d)-g(s')\tau_{\mathrm{re}}(s,d)+h(s') \big], \quad s\in\mathcal{S}_{\mathrm{re}},
\end{equation}
\end{subequations}
where $s'$, $c_\mathrm{re}(\cdot)$, and $\tau_\mathrm{re}(\cdot)$ are the next state, transition cost, and transition time defined in Section~\ref{sec:def-reform}.
The set $\mathcal{D}_{\mathrm{re}}^*(s)$ consists of the decisions attaining the minimum in Eq.~\eqref{eq:multichain-bellman-g}.

{We are now ready to develop scheduling policies based on these equations. In Section~\ref{sec:policy}, we derive an optimal index-based threshold policy for $M=2$. In Section~\ref{sec:policy-multi}, we develop policy iteration algorithms and low-complexity alternatives for the general case $M\ge2$.}

\section{Two-Modality Case: Index-Based Threshold Policy} \label{sec:policy}

\begin{figure}[t]
\centering
\includegraphics[width=0.7\linewidth]{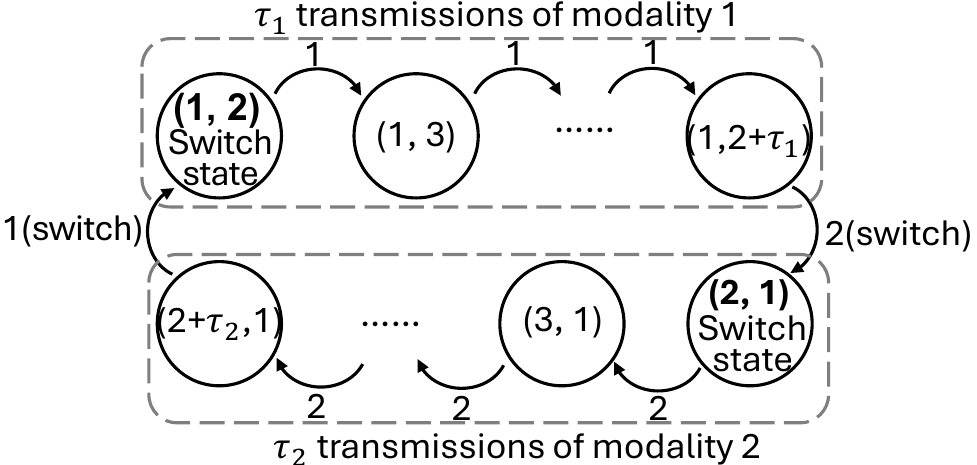}
\caption{Illustration of AoI transitions for $M=2$ with unit transmission time and zero preprocessing delay. Each circle represents an AoI state, and each arrow is labeled with the selected modality. The process has two switching states, $(1,2)$ and $(2,1)$, with reformulated decisions $\tau_1$ and $\tau_2$, respectively.}
\label{fig:AoI-transition-2}
\end{figure}

When $\mathcal{M}=\{1,2\}$, the reformulated problem reduces to a two-state SMDP, as illustrated in Fig.~\ref{fig:AoI-transition-2}.
For clarity, we restate its components below. First, by the definition of switching states in Eq.~\eqref{eq:def-switch}, the state space contains only two states:
\begin{equation}
\begin{aligned}
s_{2 \to 1}
&=
\left(
T_1^{\mathrm{pre}} + T_1^{\mathrm{tx}},\,
T_2^{\mathrm{pre}} + T_2^{\mathrm{tx}} + T_1^{\mathrm{tx}}
\right),\\
s_{1 \to 2}
&=
\left(
T_1^{\mathrm{pre}} + T_1^{\mathrm{tx}} + T_2^{\mathrm{tx}},\,
T_2^{\mathrm{pre}} + T_2^{\mathrm{tx}}
\right).
\end{aligned}
\end{equation}
The process alternates between $s_{2 \to 1}$ and $s_{1 \to 2}$ with probability~1.
Since there are only two modalities, there is no need to choose the next modality to switch to. Thus, the decision reduces to choosing the number of additional consecutive transmissions of the current modality. We use \(\tau_1,\tau_2\in\{0,\ldots,K-1\}\) to denote the decisions made at states \(s_{2 \to 1}\) and \(s_{1 \to 2}\), respectively, where $K=\max_{m\in\mathcal{M}}K_m$.
From Eq.~\eqref{eq:transition-time}, the transition time duration is $\tau_1 T^{\mathrm{tx}}_1 + T^\mathrm{tx}_2$ when starting from state $s_{2 \to 1}$ under decision $\tau_1$, and $T^\mathrm{tx}_1 + \tau_2 T^\mathrm{tx}_2$ when starting from state $s_{1 \to 2}$ under decision $\tau_2$.
Let \(c_{2\to 1}(n)\) and \(c_{1\to 2}(n)\) denote the corresponding transition costs from \(s_{2 \to 1}\) to \(s_{1 \to 2}\) and from \(s_{1 \to 2}\) to \(s_{2 \to 1}\), respectively, when the corresponding decision equals \(n\). 

Next, we present the Bellman optimality equation and characterize the optimal policy it induces in Proposition~\ref{prop:bellman-2}.

\begin{prop} \label{prop:bellman-2}
When $\mathcal{M}=\{1, 2\}$, the Bellman optimality equation for the reformulated problem~\eqref{eq:reformulated-problem} is
\begin{equation} \label{eq:bellman-2}
\begin{aligned}
    h(s_{2 \to 1}) = \min_{\tau_1} \left\{ c_{2\to 1}(\tau_1) - (\tau_1 T^\mathrm{tx}_1 + T^\mathrm{tx}_2)g + h(s_{1 \to 2}) \right\}, \\
    h(s_{1 \to 2}) = \min_{\tau_2} \left\{ c_{1 \to 2}(\tau_2) - (T^\mathrm{tx}_1 + \tau_2 T^\mathrm{tx}_2)g + h(s_{2 \to 1}) \right\}.
\end{aligned}
\end{equation}
Suppose $g$ satisfies Eq.~\eqref{eq:bellman-2}, and let $\tau_1^*$ and $\tau_2^*$ denote the optimal decisions at states $s_{2 \to 1}$ and $s_{1 \to 2}$, respectively. Then,
\begin{equation} \label{eq:optimal-policy-2}
\begin{aligned}
    \tau_1^* \in \arg\min_{\tau_1}  c_{2\to 1}(\tau_1) - \tau_1 T^\mathrm{tx}_1g , \\
    \tau_2^* \in \arg\min_{\tau_2}  c_{1 \to 2}(\tau_2) - \tau_2 T^\mathrm{tx}_2g.
\end{aligned}
\end{equation}
\end{prop}
\begin{IEEEproof}
By substituting the states, decisions, transition times, and costs for the two-modality case into the unichain Bellman optimality equation in Eq.~\eqref{eq:unichain-bellman}, we obtain Eq.~\eqref{eq:bellman-2}.
By~\cite[Theorem 3]{yushkevich1982semi}, an optimal policy attains the minimum in Eq.~\eqref{eq:bellman-2}.
Because $s_{2 \to 1}$ and $s_{1 \to 2}$ are fixed states and do not depend on the decisions $\tau_1$ and $\tau_2$, the terms $h(s_{1 \to 2})$ and $h(s_{2 \to 1})$ are constants with respect to the minimizations over $\tau_1$ and $\tau_2$, respectively.
Moreover, the term \(T^\mathrm{tx}_2 g\) is independent of \(\tau_1\), and the term \(T^\mathrm{tx}_1 g\) is independent of \(\tau_2\).
Therefore, removing these constant terms does not change the minimizers, and the optimal policy satisfies Eq.~\eqref{eq:optimal-policy-2}.
\end{IEEEproof}

From Proposition~\ref{prop:bellman-2}, Eq.~\eqref{eq:optimal-policy-2} suggests that the two optimal decisions have closed-form expressions, analogous to the analyses in~\cite{8000687,10559951,shisher2023learning,shisher2025aoi}.
Before presenting the optimal policy, for a given $\beta \in \mathbb{R}$, we define $\tau_1(\beta)$ and $\tau_2(\beta)$ as the smallest minimizers of the following problems:
\begin{equation} \label{eq:optimal-policy-beta}
\begin{aligned}
    \tau_1(\beta) \in \arg\min_{\tau_1}  c_{2\to 1}(\tau_1) - \tau_1 T^\mathrm{tx}_1\beta, \\
    \tau_2(\beta) \in \arg\min_{\tau_2}  c_{1 \to 2}(\tau_2) - \tau_2 T^\mathrm{tx}_2\beta.
\end{aligned}    
\end{equation}
For $\theta<K-1$, the index is defined with integer increments $k=1,\ldots,K-1-\theta$:
\begin{equation} \label{eq:index}
\begin{aligned}
    \gamma_1(\theta) \coloneq \min_k \frac{c_{2 \to 1}(\theta + k) - c_{2 \to 1}(\theta)}{kT^\mathrm{tx}_1}, \\
    \gamma_2(\theta) \coloneq \min_k \frac{c_{1 \to 2}(\theta + k) - c_{1 \to 2}(\theta)}{kT^\mathrm{tx}_2}, \\
\end{aligned}
\end{equation}
where the numerator is the additional cumulative cost and the denominator is the additional transition duration, both when the decision is $\theta + k$ instead of $\theta$ (see Remark~\ref{remark:1} in this section for a detailed explanation of the index function).

\begin{theorem}\label{thm:main}
Suppose $g$ satisfies Eq.~\eqref{eq:bellman-2}, and choose the smallest optimal decisions $\tau_1^*$ and $\tau_2^*$ for the reformulated problem~\eqref{eq:reformulated-problem} at states $s_{2 \to 1}$ and $s_{1 \to 2}$, respectively.
The following assertions are true:
\begin{enumerate}[label=(\roman*), itemindent=0pt]
    \item For $m = 1, 2$, the solution to Eq.~\eqref{eq:optimal-policy-beta} is
    \begin{equation} \label{eq:opt-beta}
        \tau_m(\beta) = \min \left\{ \theta : \gamma_m(\theta) \ge \beta \right\},
    \end{equation}
    where $\gamma_m(\cdot)$ is defined in Eq.~\eqref{eq:index}.
    \item The optimal decisions $\tau_1^*$ and $\tau_2^*$ are
    \begin{equation} \label{eq:index-main}
        \tau_m^* = \tau_m(g) = \min \left\{ \theta : \gamma_m(\theta) \ge g \right\}.
    \end{equation}
\item The value of $g$ is the unique root of
\begin{equation} \label{eq:Lopt}
    f(\beta) = 0,
\end{equation}
where $f(\beta)$ is defined as
\begin{equation} \label{eq:f}
\begin{aligned}
    f(\beta) \coloneq\;& c_{2 \to 1}(\tau_1(\beta)) + c_{1 \to 2}(\tau_2(\beta)) \\
    &- \beta \bigl((\tau_1(\beta) + 1)T^\mathrm{tx}_1 + (\tau_2(\beta) + 1)T^\mathrm{tx}_2\bigr).
\end{aligned}
\end{equation}
\end{enumerate}
\end{theorem}

\begin{proofsketch}
We prove assertion~(\romannumeral 1) by induction.
Specifically, we first show that \(\tau_m(\beta) = 0\) if $\min \left\{ \theta : \gamma_m(\theta) \ge \beta \right\} = 0$. Then, if $\min \left\{ \theta : \gamma_m(\theta) \ge \beta \right\} = j$, it is equivalent to two conditions:
(a) \(\gamma_m(i) < \beta\) for all \(i=0,1,\dots,j-1\), and
(b) \(\gamma_m(j) \ge \beta\).
From condition (a), we show that \(\tau_m(\beta) \ge j\); from condition (b), we show that \(\tau_m(\beta) \le j\).
Therefore, we conclude that \(\tau_m(\beta) = j\), which concludes the induction.
By Proposition~\ref{prop:bellman-2} and assertion~(\romannumeral 1), we obtain assertion~(\romannumeral 2).
By substituting the decisions $\tau_m(\beta)$ into the Bellman optimality equation~\eqref{eq:bellman-2} and solving for $g$ and $h(\cdot)$, we obtain the result in assertion~(\romannumeral 3). We provide the complete proof in Appendices~\ref{sec:proof-main} and~\ref{sec:prop-tau}.
\end{proofsketch}

Eq.~\eqref{eq:index-main} in assertion~(\romannumeral 2) suggests that the optimal stationary policy exhibits an \emph{index-based threshold} structure.
That is, the policy selects modality $m$ for $\theta$ additional consecutive transmissions, stopping at the first time when the index function $\gamma_m(\theta)$ reaches the threshold $g$; then, the policy switches to select the other modality. One interesting observation is that the two modalities share the same threshold $g$, which is exactly the optimal average cost $\bar{L}_{\mathrm{opt}}$.
The index function of each modality can be pre-computed independently, as it depends only on the known parameters (i.e., the inference error function, the preprocessing times, and the transmission times).
The last assertion shows that the threshold $g$ can be determined by solving Eq.~\eqref{eq:Lopt}.
The following proposition shows that Eq.~\eqref{eq:Lopt} can be efficiently solved.
\begin{prop} \label{prop:g}
     The function $f$ defined in Eq.~\eqref{eq:f} satisfies:
    \begin{enumerate}[label=(\roman*), labelindent=0pt]
        \item It is concave, continuous, and strictly decreasing,
        \item $\lim_{\beta \to \infty} f(\beta) = -\infty$ and $\lim_{\beta \to -\infty} f(\beta) = \infty$.
    \end{enumerate}    
\end{prop}
\begin{IEEEproof}
    The function $f$ is the minimum of finitely many affine functions of $\beta$, each with a strictly negative slope. The complete proof is in~\cite[Lemma 9]{10559951}.
\end{IEEEproof}
Given these properties of $f$, we can efficiently solve Eq.~\eqref{eq:Lopt} (i.e., $f(\beta)=0$) using low-complexity algorithms such as bisection search and  Newton’s method~\cite[Algorithms 1-3]{9437348}.

\begin{remark} \label{remark:prior-two-modality}
A similar result for two-modality scheduling was obtained independently in~\cite{shisher2025aoi}. Our result accounts for the asymmetric setting with $T^{\mathrm{tx}}_1 \neq T^{\mathrm{tx}}_2$, whereas~\cite{shisher2025aoi} assumes unit transmission times for both sources. Moreover, we show that the two-modality case ($M=2$) has a fundamentally different chain structure from that of the case $M\ge 3$. This difference between $M=2$ and $M\ge 3$ was not discussed in~\cite{shisher2025aoi}.
\end{remark}


\begin{remark} \label{remark:1}
The index function $\gamma_m(\theta)$ reflects the \textbf{minimum future cost} if the scheduler continues to select modality $m$ after having selected it for $\theta$ additional consecutive transmissions since the last switching time.
To see this, consider a case with unit transmission times and no preprocessing delay.
For $\theta<K-1$, the index function of modality $1$ becomes
\begin{equation*}
    \gamma_1(\theta) = \min_k \frac{\sum_{j=1}^{k} L(1 , \theta + 2 + j)}{k}.
\end{equation*}
After $\theta$ additional transmissions of modality $1$, the process reaches state $(1, \theta + 2)$. Function $\gamma_1(\theta)$ is the minimum average cost starting from state $(1, \theta + 2)$ until switching after $k$ more transmissions.

\end{remark}

\begin{remark} \label{remark:2}
Due to the non-monotonic AoI functions, our index function is not necessarily monotonic. This result generalizes the two-source scheduling problem in remote estimation~\cite{6126061}, where the estimation error is a monotonic function of AoI.
     Specifically,  consider a case with unit transmission times and no preprocessing delay.
     When the inference error is a non-decreasing function of AoI, the index function of modality 1 for $\theta<K-1$ reduces to
    \begin{equation*}
        \gamma_1(\theta) = \min_k \frac{\sum_{j=\theta + 2}^{\theta + k + 1}L (1, j + 1)}{k} = L (1, \theta + 3),
    \end{equation*}
    where the last equality holds as the minimum is achieved when $k=1$ and $L$ is non-decreasing.
    Then, the policy switches at the first $\theta<K-1$ satisfying $L(1,\theta+3)\ge\bar{L}_{\mathrm{opt}}$, or at $\theta=K-1$ otherwise. This has the threshold structure in~\cite[Proposition 3.5]{6126061}, with the finite decision bound imposed here.
\end{remark}

\section{General-Modality Case} \label{sec:policy-multi}
{We now develop scheduling policies for the general case when $M\ge2$.
From the original problem, we obtain a policy iteration algorithm that considers at most $K^M$ states and $M$ decisions per state, giving a per-iteration time complexity of $\mathcal{O}(MK^M)$ with precomputed transition data.
The resulting policy requires $\mathcal{O}(K^M)$ parameters, since it selects one modality at each state.
We next show that using our reformulated problem yields a policy iteration algorithm with fewer policy parameters and lower per-iteration time complexity.}

\subsection{Optimal Switching-Type Policy (OPT-ST)} \label{subsec:pi-M}
Recall that at each decision epoch, the optimal policy jointly determines the number of consecutive transmissions of the current modality (i.e., transmission) and the next modality to switch to (i.e., switching).
{We compute OPT-ST by solving the Bellman optimality equation~\eqref{eq:Bellman-equation}.
For $M>2$, the next switching state depends on the decision, and a closed-form solution remains unknown.
To solve the equation for general $M\ge2$, we use multichain policy iteration to compute an \emph{optimal} stationary policy, as presented in Alg.~\ref{alg:mcpi}.}

{For computation, we search over $d_\tau\in\{0,\ldots,K-1\}$ and $d_s\in\mathcal{M}\setminus\{n\}$ when the current modality is $n$.
The decision sets used in Alg.~\ref{alg:mcpi} are restricted accordingly.}
\begin{algorithm}[!t]
\caption{Multichain Policy Iteration Algorithm}
\label{alg:mcpi}
\algrenewcommand\algorithmicrequire{\textbf{Input:}}
\algrenewcommand\algorithmicensure{\textbf{Output:}}
\begin{algorithmic}[1]
\Require Number of modalities $M$; inference error function $L$; preprocessing delay $T_m^{\mathrm{pre}}$; transmission time $T_m^{\mathrm{tx}}$; truncation vector $\boldsymbol{K}$.
\Ensure A stationary policy $\pi_{\mathrm{re}}^*$.
\State {Arbitrarily assign a unique index to each state.} \label{line:indexing}
\State Initialize a stationary policy $\pi^{(0)}_\mathrm{re}$ randomly.
\For{$i=0,1,2,\ldots$}
    \State Step 1 \textbf{(Policy Evaluation)}.
    \State Find all recurrent classes induced by the policy $\pi_\mathrm{re}^{(i)}$. \label{line:recurrent-classes}
    \For{each recurrent class}
        \State Solve $g^{(i)}(s)$ using Eq.~\eqref{eq:gu}.
    \EndFor
    \State $s_{\mathrm{ref},j}^{(i)}$: state in $\mathcal{R}_j^{(i)}$ with smallest index.
    \State Solve $h^{(i)}(s)$ using Eq.~\eqref{eq:h}.
    \State Step 2 \textbf{(Policy Improvement)}.
    \For{each $s \in \mathcal{S}_\mathrm{re}$}
        \State $\Gamma_s^{(i)} \gets \{d \in \mathcal{D}_{\mathrm{re}}(s) \ | \ d \ \text{minimizes} \ g^{(i)}(s')\}$.
        \State $\Lambda_s^{(i)} \gets \{d \in \Gamma_s^{(i)} \ | \ d \ \text{minimizes Eq.~\eqref{eq:multichain-bellman-h}} \}$.
        \If{$\pi_\mathrm{re}^{(i)}(s) \in \Lambda_s^{(i)}$}
            \State $\pi_\mathrm{re}^{(i+1)}(s) \gets \pi_\mathrm{re}^{(i)}(s)$.
        \Else
            \State $\pi_\mathrm{re}^{(i+1)}(s) \gets \ \text{arbitrary} \ d \in \Lambda_s^{(i)}$.
\EndIf
    \EndFor
    \State \textbf{If} $\pi_\mathrm{re}^{(i+1)}=\pi_\mathrm{re}^{(i)}$, \textbf{return} $\pi_{\mathrm{re}}^* \gets \pi_\mathrm{re}^{(i)}$. \Comment{termination}

\EndFor
\end{algorithmic}
\end{algorithm}


The algorithm proceeds by iteratively evaluating the current policy and improving it until the policy converges.
We highlight the key differences between multichain and unichain policy iteration, as well as the modifications tailored to our problem.
{In the policy evaluation step, at iteration $i$, the policy $\pi_{\mathrm{re}}^{(i)}$ induces a Markov chain that may contain multiple recurrent classes, denoted by $\mathcal{R}_1^{(i)},\ldots,\mathcal{R}_{k_i}^{(i)}$.
Unlike in the unichain case, $g(\cdot)$ and $h(\cdot)$ must be determined for each recurrent class.
Furthermore, as both $\pi_{\mathrm{re}}^{(i)}$ and the state transitions are deterministic, each transient state eventually enters a unique recurrent class. Let $\mathcal{T}_j^{(i)}$ denote the set of transient states that eventually enter $\mathcal{R}_j^{(i)}$.}
We can use the Fox-Landi algorithm~\cite{fox1968scientific} to identify recurrent and transient states.
{The average cost is constant within each recurrent class and its associated transient states. For $j=1,\ldots,k_i$,
\begin{equation} \label{eq:gu}
            g^{(i)}(u) = \frac{\sum_{s \in \mathcal{R}^{(i)}_j} c_\mathrm{re}(s,\pi^{(i)}_\mathrm{re}(s))}{\sum_{s \in \mathcal{R}^{(i)}_j} \tau_{\mathrm{re}}(s, \pi^{(i)}_\mathrm{re}(s))}, \ \forall u\in \mathcal{R}^{(i)}_j \cup \mathcal{T}^{(i)}_j.
\end{equation}
}

Although $g(\cdot)$ is uniquely determined within each recurrent class, the relative value function \( h(\cdot) \) is not unique but is determined only up to an additive constant~\cite{schweitzer1978foolproof}.
To determine the relative value function, we follow the suggestion in~\cite{schweitzer1978foolproof}: we assign an index to each state and always select the state with the smallest index in each recurrent class as the reference state.
That is, we set $h^{(i)}(s_{\mathrm{ref},j}^{(i)}) = 0$ if $s_{\mathrm{ref},j}^{(i)}$ has the smallest index in class $\mathcal{R}_j^{(i)}$.
For the other states $s$, we have
\begin{equation} \label{eq:h}
    h^{(i)}(s) = c_\mathrm{re}(s,\pi_\mathrm{re}^{(i)}(s)) - g^{(i)}(s)\tau_\mathrm{re}(s,\pi_\mathrm{re}^{(i)}(s)) + h^{(i)}(s').
\end{equation}
{This index is assigned arbitrarily at the beginning and remains fixed throughout the iteration.}
Note that an arbitrary choice of the reference state may cause policy iteration to fail to converge (see Example 1 in~\cite{schweitzer1978foolproof}).

The policy improvement step for the multichain model differs from that of the unichain model because the multichain Bellman optimality equation includes the additional optimality equation Eq.~\eqref{eq:multichain-bellman-g}.
Accordingly, the algorithm first selects decisions that lead the state to the recurrent class with the smallest average cost, according to the first optimality equation Eq.~\eqref{eq:multichain-bellman-g}.
If multiple decisions satisfy this condition, the algorithm then selects the one that yields the smallest relative value function, following the second optimality equation Eq.~\eqref{eq:multichain-bellman-h}.
Finally, the algorithm terminates when no decision can be further improved, that is, when the policy has converged.

{\textbf{Complexity analysis.}
The detailed complexity analysis of policy iteration and coordinate descent is provided in Appendix~\ref{app:complexity}.
By using the reformulated problem, we reduce the state set to at most $M(M-1)K^{M-2}$ states, according to Eq.~\eqref{eq:reformulated-state}.
The algorithm considers at most $(M-1)K$ decisions per state.
Policy iteration therefore has per-iteration complexity $\mathcal{O}\!\left(M(M-1)^2K^{M-1}\right)$.
Compared with $\mathcal{O}(MK^M)$ for the original problem, it reduces the exponent of $K$ from $M$ to $M-1$.
The ratio of the original complexity expression to the reformulated one is $K/(M-1)^2$.
For example, when $K=100$ and $M=3$, this ratio is $25$.
Similarly, as each state requires two decisions, the policy can be represented by $\mathcal{O}\!\left(M(M-1)K^{M-2}\right)$ parameters, compared with $\mathcal{O}(K^M)$ for the original problem.}

\subsection{Round-Robin-Like Switching-Type Policy (RR-ST)} \label{sec:low-complexity}
{Our optimal algorithm, OPT-ST, searches over both the transmission number and the next modality at every switching state.
To further reduce computational complexity, we fix the switching order, which reduces both the state set and the decision set.}


We first consider a setting in which the scheduler follows a fixed cyclic order. Without loss of generality, we write this order as $(1,2,\dots,M)$, because the indices can represent any modalities. We assume that this order is given to the scheduler as an oracle; in practice, such an order can arise from cyclic polling-based systems~\cite{vishnevsky2021polling}.

Under the fixed cyclic order, the decision at each switching state is simplified from $d_{\mathrm{re}}=(d_\tau,d_s)$ to $d_\tau$, since the next modality is predetermined by the cyclic order. All other components of the reformulated SMDP remain unchanged. Therefore, we can directly apply the MPI procedure in Alg.~\ref{alg:mcpi} to obtain the RR-ST policy.

{\textbf{Complexity analysis.}
Fixing the switching order simplifies both the state set and the decisions at each state.
Each modality has a fixed predecessor in the cycle, leaving only $M$ switching pairs.
Since two AoI values are known for each pair, the remaining $M-2$ AoI values give at most $K^{M-2}$ states per pair and $MK^{M-2}$ states in total.
With the next modality fixed as well, the scheduler only chooses the number of additional transmissions, giving $K$ candidate decisions per state.
RR-ST therefore has a per-iteration time complexity of $\mathcal{O}(MK^{M-1})$.
The ratio of the complexity expression for OPT-ST to that for RR-ST is $(M-1)^2$.
Furthermore, representing the policy requires only one transmission number per state, or $\mathcal{O}(MK^{M-2})$ parameters, compared with $\mathcal{O}\!\left(M(M-1)K^{M-2}\right)$ for OPT-ST.}

\begin{algorithm}[!t]
\caption{Coordinate Descent Algorithm}
\label{alg:foft-cd}
\algrenewcommand\algorithmicrequire{\textbf{Input:}}
\algrenewcommand\algorithmicensure{\textbf{Output:}}
\begin{algorithmic}[1]
\Require Number of modalities $M$; inference error function $L$; preprocessing delay $T_m^{\mathrm{pre}}$; transmission time $T_m^{\mathrm{tx}}$; truncation vector $\boldsymbol{K}$.
\Ensure A decision vector \(\boldsymbol{\tau}\).
\State Initialize \(\boldsymbol{\tau}^{(0)}\) randomly.
\For{$i = 1,2,\ldots$}
    \State \(\boldsymbol{\tau}^{(i)} \gets \boldsymbol{\tau}^{(i-1)}\).
    \For{$m = 1,2,\dots,M$}
        \State Update \(\tau_m^{(i)}\) by Eq.~\eqref{eq:cd}.
    \EndFor
    \If{$\boldsymbol{\tau}^{(i)} = \boldsymbol{\tau}^{(i-1)}$}
        \State \textbf{return} \(\boldsymbol{\tau}^{(i)}\). \Comment{termination}
    \EndIf
\EndFor
\end{algorithmic}
\end{algorithm}

\begin{table}[!t]
\centering
\caption{Summary of Complexity Comparison}
\label{tab:complexity-summary}
\setlength{\tabcolsep}{3pt}
\begin{tabular*}{\columnwidth}{@{\extracolsep{\fill}}lcc@{}}
\hline
Algorithm & \makecell{Time complexity\\(per-iteration)} & \makecell{Required\\parameters} \\
\hline
\makecell[l]{Policy iteration\\(original problem)} & $\mathcal{O}(MK^M)$ & $K^M$ \\
OPT-ST & $\mathcal{O}\!\left(M(M-1)^2K^{M-1}\right)$ & $2M(M-1)K^{M-2}$ \\
RR-ST & $\mathcal{O}(MK^{M-1})$ & $MK^{M-2}$ \\
RR-Thres & $\mathcal{O}(MK)$ & $M$ \\
\hline
\end{tabular*}
\end{table}

\subsection{Round-Robin-Like Threshold Policy (RR-Thres)}
{Although OPT-ST and RR-ST reduce computational complexity, their per-iteration complexity still grows exponentially with $M$ because they optimize a decision at every switching state. To this end, we consider assigning a fixed decision to each modality. This leads to a round-robin-like threshold policy.} Specifically, the switching order \((1,2,\dots,M)\) is fixed, and whenever the scheduler switches to modality \(m\), it transmits that modality for \(\tau_m \in \{0,\ldots,K-1\}\) extra consecutive times. Let \(\boldsymbol{\tau} \coloneq (\tau_1,\dots,\tau_M)\) denote the decision variable.
Proposition~\ref{prop:foft-state} shows that, under any decision \(\boldsymbol{\tau}\), the process cycles among the \(M\) states.

\begin{prop}\label{prop:foft-state}
Given any decision \(\boldsymbol{\tau}\), the process cycles among the \(M\) switching states
\[
s_{1\to 2}(\boldsymbol{\tau}), \dots, s_{M-1\to M}(\boldsymbol{\tau}), s_{M\to 1}(\boldsymbol{\tau}),
\]
which are uniquely determined by \(\boldsymbol{\tau}\). Furthermore, the reformulated problem~\eqref{eq:reformulated-problem}, restricted to round-robin-like threshold policies (RR-Thres), reduces to
\begin{equation} \label{eq:foft-problem}
\min_{\boldsymbol{\tau}} J(\boldsymbol{\tau}) \coloneq
\frac{
\sum_{i=1}^{M}
c_{\mathrm{re}}\!\left(
s_{(i-1)\to i}(\boldsymbol{\tau}),
\tau_{i}
\right)
}{
\sum_{i=1}^{M}(\tau_{i} + 1)T^{\mathrm{tx}}_{i}
},
\end{equation}
where $s_{0 \to 1}(\boldsymbol{\tau}) = s_{M \to 1}(\boldsymbol{\tau})$.
\end{prop}
\begin{proofsketch}
As the state evolution is deterministic, each switching state can be written as a function of the threshold vector \(\boldsymbol{\tau}\). Consequently, under any policy $\boldsymbol{\tau}$, the system eventually cycles among \(M\) switching states. The time-average cost is therefore equal to the total cost incurred over one cycle divided by the corresponding cycle duration, i.e., Eq.~\eqref{eq:foft-problem}. See Appendix~\ref{app:CD} for the complete proof.
\end{proofsketch}

Eq.~\eqref{eq:foft-problem} in Proposition~\ref{prop:foft-state} shows that the process is uniquely determined by the decision vector \(\boldsymbol{\tau}\).
This property enables a simple heuristic coordinate descent algorithm, as given in Alg.~\ref{alg:foft-cd}.
The algorithm updates \(\boldsymbol{\tau}\) one coordinate at a time: in the $i$-th step, it optimizes \(\tau^{(i)}_m\) while keeping the other coordinates fixed, i.e.,
\begin{equation}\label{eq:cd}
\tau_m^{(i)}
\in \arg\min_{u}
J\!\left(
\tau_1^{(i)}, \ldots, \tau_{m-1}^{(i)},
u,
\tau_{m+1}^{(i-1)}, \ldots, \tau_M^{(i-1)}
\right).
\end{equation}
To break ties, the current decision remains unchanged if it attains the minimum; otherwise, an arbitrary minimizer is selected.
{This procedure is repeated until \(\boldsymbol{\tau}\) converges.
Because ties retain the current decision, every change strictly decreases $J$.
The finite search grid therefore guarantees termination at a coordinatewise minimum on this grid, although this heuristic need not find its global minimum.}


{\textbf{Complexity analysis.}
Each sweep of coordinate descent updates the $M$ thresholds one at a time and checks at most $K$ candidate values for each threshold.
Thus, RR-Thres has a per-iteration time complexity of $\mathcal{O}(MK)$.
The policy requires only $M$ parameters, one threshold per modality.}

{Table~\ref{tab:complexity-summary} summarizes the per-iteration complexity and policy parameter bounds of the four algorithms.}
Specifically, when \(K=50\) and \(M=4\), the upper bounds on the numbers of parameters are \(6{,}250{,}000\), \(60{,}000\), \(10{,}000\), and \(4\), respectively.

\begin{remark}
We use an example to illustrate the difference between the round-robin-like switching-type policy and the round-robin-like threshold policy. Consider $M=3$, with unit transmission time and no preprocessing delay. Under a policy in the fixed-order setting, the state may evolve as
\begin{multline*}
(1,2,3)\xrightarrow{3}(2,3,1)\xrightarrow{2}(3,1,2)\xrightarrow{2}(4,1,3)\xrightarrow{1}(1,2,4)\\
\xrightarrow{1}(1,3,5)\xrightarrow{3}(2,4,1)\xrightarrow{3}(3,5,1)\xrightarrow{2}(4,1,2)\xrightarrow{1}(1,2,3),
\end{multline*}
where the label on each arrow denotes the selected modality. This example shows that, without a fixed threshold for each modality, the process may cycle through more than $M$ states.
\end{remark}

\section{NUMERICAL RESULTS} \label{sec:numerical}

\begin{figure*}[t]
    \centering    
    \begin{subfigure}[t]{0.32\textwidth}
        \centering
        \includegraphics[width=\linewidth]{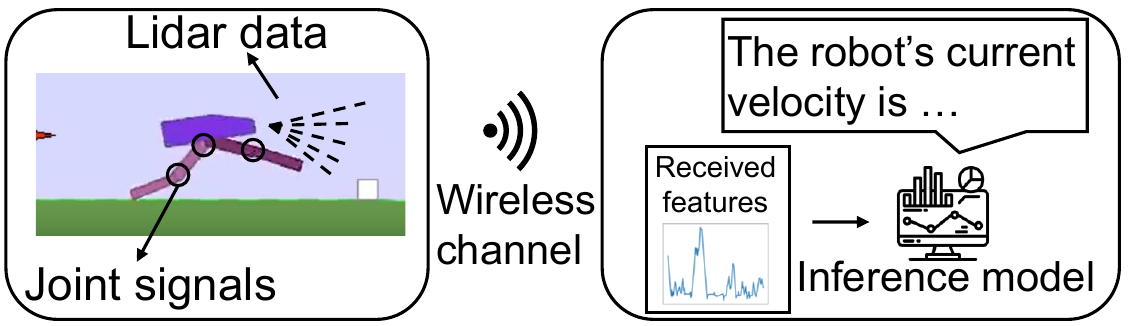}
        \caption{Robot state prediction.}
        \label{fig:setup_case1}
    \end{subfigure}
    \hfill
    \begin{subfigure}[t]{0.32\textwidth}
        \centering
        \includegraphics[width=\linewidth]{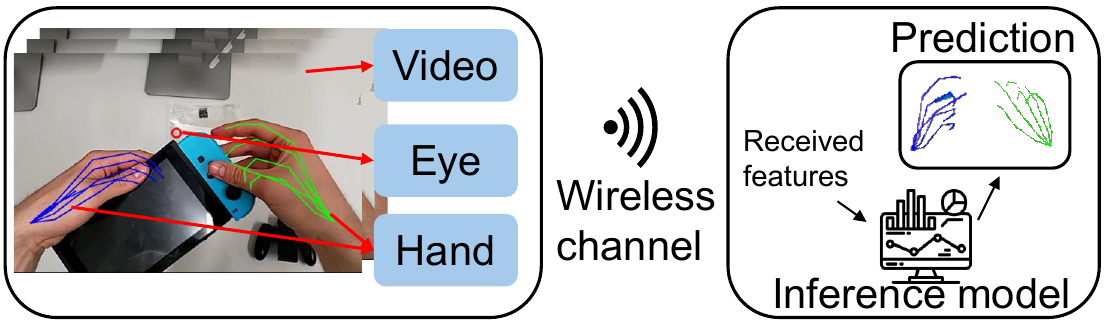}
        \caption{Hand pose prediction.}
        \label{fig:setup_case2}
    \end{subfigure}
    \hfill
    \begin{subfigure}[t]{0.32\textwidth}
        \centering
        \includegraphics[width=\linewidth]{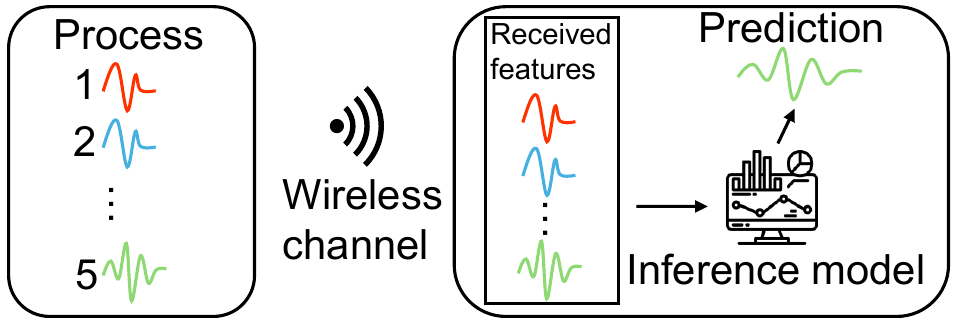}
        \caption{Vector autoregressive process prediction.}
        \label{fig:setup_case3}
    \end{subfigure}
    \caption{Experimental setup under three different cases.}
    \label{fig:setup_all}
\end{figure*}

\begin{figure*}[!t]
\centering

\begin{subfigure}[t]{0.28\textwidth}
    \centering
    \includegraphics[width=\textwidth]{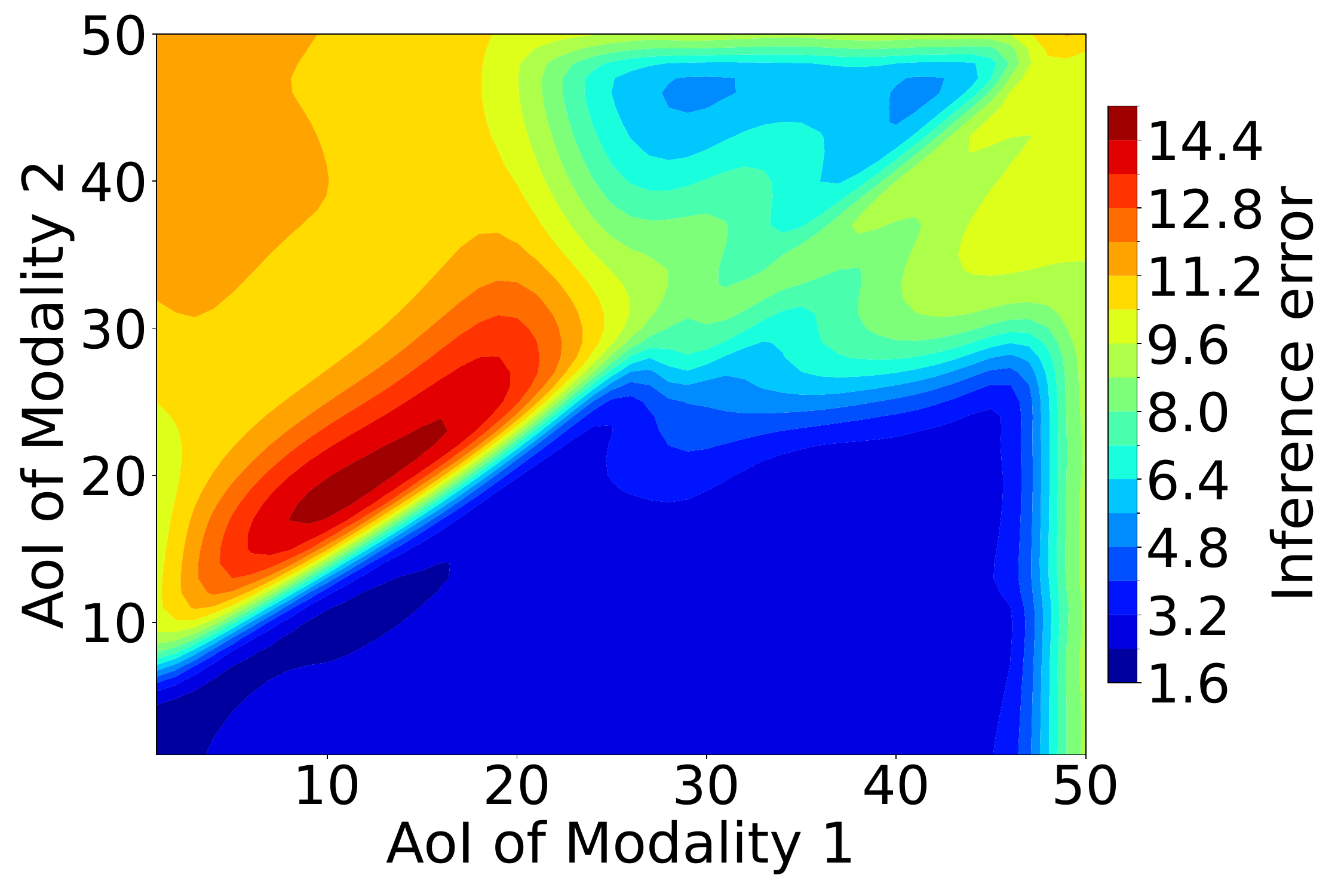}
    \caption{Robot state prediction (Modality 1: control signals; Modality 2: LiDAR).}
    \label{fig:2D}
\end{subfigure}
\hfill
\begin{subfigure}[t]{0.26\textwidth}
    \centering
    \includegraphics[width=\textwidth]{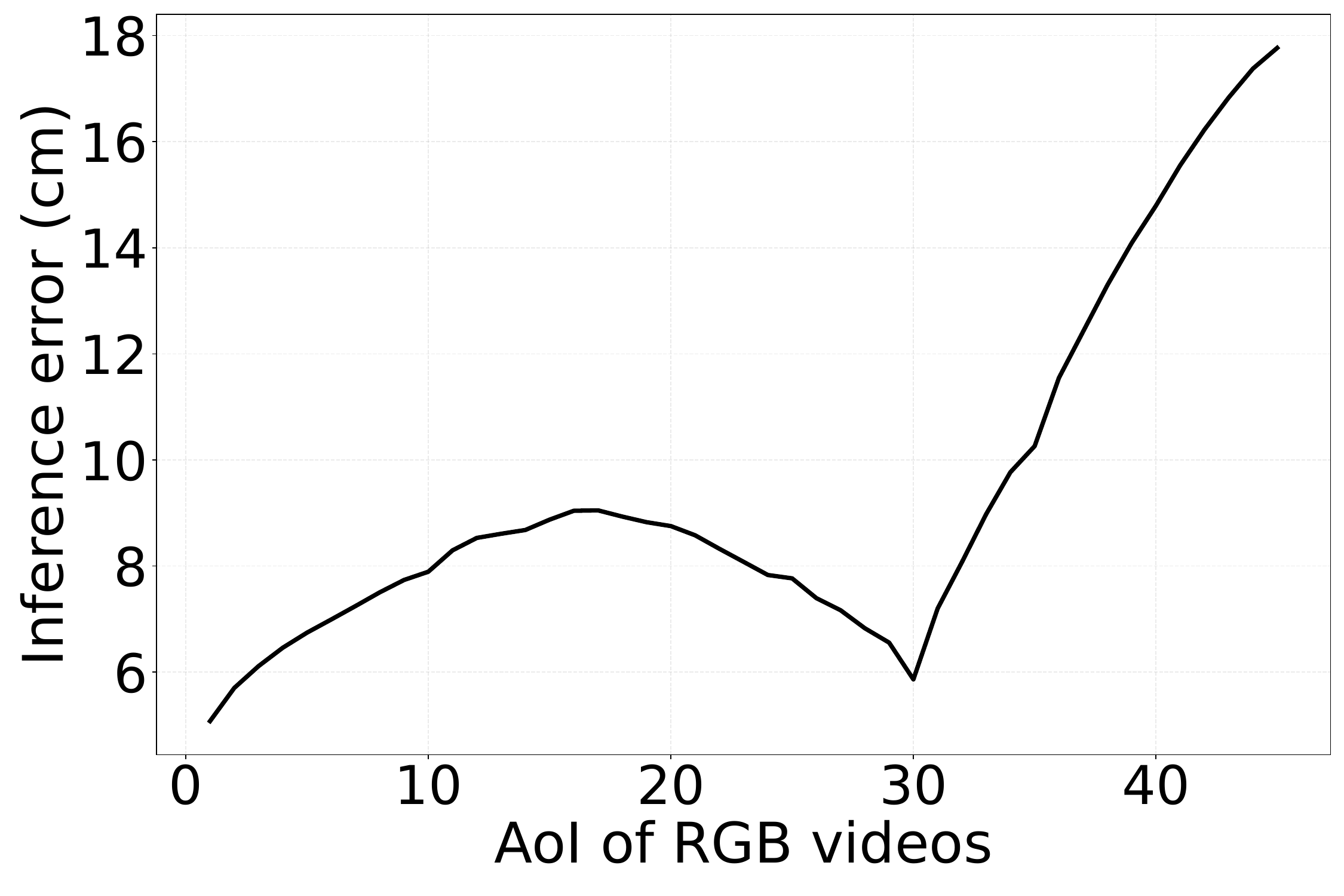}
    \caption{Hand pose prediction. The AoIs of eye gaze and hand motion are 30.}
    \label{fig:3D}
\end{subfigure}
\hfill
\begin{subfigure}[t]{0.26\textwidth}
    \centering
    \includegraphics[width=\textwidth]{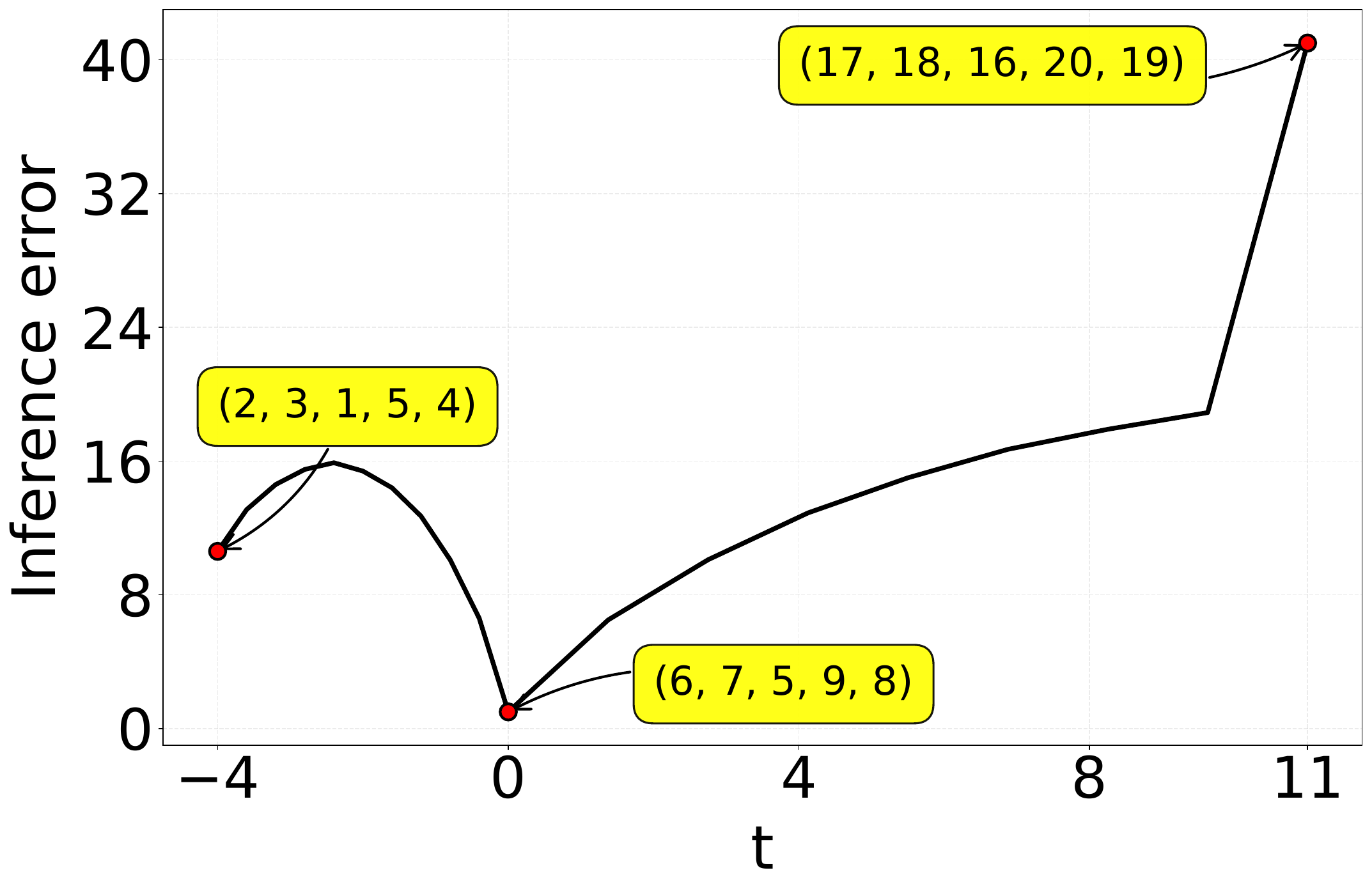}
    \caption{Vector autoregressive process prediction. The curve shows the inference error under the AoI vectors $\boldsymbol{\nu} + t\mathbf{1}$ when $\boldsymbol{\nu} = (6,7,5,9,8)$.}
    \label{fig:5D}
\end{subfigure}

\caption{Inference error under different AoI vectors.}
\label{fig:inference error}

\end{figure*}

In this section, we present numerical results from three case studies with different numbers of modalities to evaluate our multimodal remote inference system.
All experiments were run on a server with an AMD EPYC 7313 CPU (16 cores) and a single NVIDIA A2 GPU.

{
\subsection{Obtaining the Inference Error for Data-Driven Models}

The training and evaluation of a data-driven inference model typically involve three processes: training, validation, and testing.
In particular, validation evaluates the predictor’s performance on a separate validation set during training and helps determine when to stop training.
In our setting, the inference error function is obtained during model validation.

Specifically, given a dataset, we first split it into training, validation, and test sets containing 60\%, 20\%, and 20\% of the data, respectively.
To account for different AoI vectors, we augment each set as follows.
For each original feature-label pair $(\mathbf{X}_t,Y_t)$ and each AoI vector $\boldsymbol{\delta}\in\mathcal{K}$, we construct an augmented sample $(\mathbf{X}_{t-\boldsymbol{\delta}},\boldsymbol{\delta};Y_t)$, where $\mathbf{X}_{t-\boldsymbol{\delta}}=(X_{1,t-\delta_1},\ldots,X_{M,t-\delta_M})$ contains historical features with the corresponding ages.
We use the augmented training set to train a single predictor that handles different AoI vectors.

During model validation, we compute the inference loss for each augmented validation sample.
For each AoI vector $\boldsymbol{\delta}$, we average the losses over all validation samples constructed with that vector to obtain the corresponding empirical inference error $L(\boldsymbol{\delta})$.
Obtaining the inference error function requires only an additional averaging operation for each AoI vector, as the loss for each sample is already computed during validation.
Therefore, the additional computational cost is small compared with the overall cost of model validation. 
}

{

\subsection{Application Scenarios}
We consider three applications: (\romannumeral 1) a simulation environment in the two-modality case, (\romannumeral 2) real-life trace data in the three-modality case, and (\romannumeral 3) synthetic data in the five-modality case. We illustrate each case in Fig.~\ref{fig:setup_all}.

\textbf{Two-Modality Case: Simulated Robot State Prediction.}
We consider a robot performing a task at a remote location and monitor it by inferring its current velocity using two modalities: LiDAR measurements and joint control signals.
We generate a time-series dataset using the Bipedal Walker simulation environment~\cite{towers2024gymnasium} and train a long short-term memory (LSTM) network for this task.
We use mean squared error (MSE) as the inference loss.
Details of data generation and model setup are provided in Appendix~\ref{app:experimental-setup}.

\textbf{Three-Modality Case: Hand Pose Prediction.}
We consider a hand pose prediction task in which we predict future hand joint positions using three modalities: RGB video, hand motion, and eye gaze.
We use multimodal data from the HoloAssist dataset~\cite{wang2023holoassist}, collected by an augmented reality (AR) device, and adopt a Seq2Seq model~\cite{sutskever2014sequence} for prediction.
We use the Euclidean distance between the predicted and true joint positions as the inference loss, measured in centimeters.
Details of data preparation and model setup are provided in Appendix~\ref{app:experimental-setup}.

\begin{table*}[t]
\centering
\caption{Comparison of scheduling algorithms under different scenarios.}
\label{tab:cost_time_all}

\setlength{\tabcolsep}{3pt}

\begin{tabular}{lcccc|cccc|cccc}
\hline

& \multicolumn{4}{c|}{Robot state prediction}
& \multicolumn{4}{c|}{Hand pose prediction}
& \multicolumn{4}{c}{Vector autoregressive process prediction} \\
\cline{2-5}
\cline{6-9}
\cline{10-13}

Algorithm
& MSE
& \shortstack{Offline\\time (s)}
& \shortstack{Storage\\overhead}
& \shortstack{Online\\time ($\mu$s)}
& \shortstack{$L_2$ error\\(cm)}
& \shortstack{Offline\\time (s)}
& \shortstack{Storage\\overhead}
& \shortstack{Online\\time ($\mu$s)}
& MSE
& \shortstack{Offline\\time (s)}
& \shortstack{Storage\\overhead}
& \shortstack{Online\\time ($\mu$s)} \\
\hline

Round-robin
& 11.4 & 0 & 0 B & 0.534
& 3.21 & 0 & 0 B & 0.622
& 13.1 & 0 & 0 B & 0.476 \\

Uniform random
& 7.59 & 0 & 0 B & 19.6
& 5.51 & 0 & 0 B & 20.0
& 20.2 & 0 & 0 B & 20.3 \\

Greedy
& 8.94 & 0 & 0 B & 38.0
& 4.57 & 0 & 0 B & 104
& 41.1 & 0 & 0 B & 46.1 \\

RL
& \textbf{4.50} & 34.6 & 24.0 B & 2.25
& \textbf{1.64} & 370 & 582 B & 2.78
& 10.1 & $1.70\times10^{3}$ & 2.44 MiB & 3.96 \\

\hline

Index (ours)
& \textbf{4.50} & $2.24\times10^{-3}$ & 2.00 B & 0.914
& -- & -- & -- & --
& -- & -- & -- & -- \\

OPT-ST (ours)
& \textbf{4.50} & $7.81\times10^{-3}$ & 24.0 B & 2.34
& \textbf{1.64} & 0.616 & 582 B & 2.86
& \textbf{8.08} & 22.1 & 2.44 MiB & 4.20 \\

RR-ST (ours)
& \textbf{4.50} & $7.92\times10^{-3}$ & 24.0 B & 2.07
& 2.79 & 0.461 & 315 B & 3.25
& 9.96 & 4.21 & 1.22 MiB & 4.09 \\

RR-Thres (ours)
& \textbf{4.50} & $8.58\times10^{-3}$ & 4.00 B & 0.868
& 2.79 & 0.129 & 6.00 B & 0.873
& 11.2 & 0.246 & 10.0 B & 0.798 \\

\hline
\end{tabular}
\end{table*}
\textbf{Five-Modality Case: Autoregressive Process Prediction.}
We generate an $M$-dimensional $(M=5)$ vector autoregressive (VAR) process,
\[
    \boldsymbol{X}_t
    = B_1\boldsymbol{X}_{t-1}+\cdots+B_p\boldsymbol{X}_{t-p}
    +\boldsymbol{u}_t,
    \quad t=0,1,\ldots,
\]
where $p$ is the model order, $B_i$ are coefficient matrices, and $\boldsymbol{u}_t$ is white noise.
For $m=1,\ldots,M-1$, each modality follows a first-order AR process:
\begin{equation}
    X_{m,t}=\rho_m X_{m,t-1}+u_{m,t},
    \quad t=0,1,\ldots.
\end{equation}
The last modality depends on delayed values of all modalities:
\begin{equation}\label{eq:xMt}
    X_{M,t}=\sum_{m=1}^{M} b_m X_{m,t-\nu_m}+u_{M,t},
    \quad t=0,1,\ldots,
\end{equation}
where $b_m$ is a coefficient and $\nu_m$ is the time lag associated with modality $m$.
Intuitively, the received signal of modality $m$ is particularly informative for predicting $X_{M,t}$ when its AoI is $\nu_m$.
We set $p=10$, $\rho_m=0.95$, and $b_m=0.95$, and randomly select each $\nu_m\in\{1,\ldots,p\}$.
Each $u_{m,t}$ is zero-mean Gaussian noise with standard deviation $0.005$.

In contrast to the data-driven inference models, we derive the inference error function directly, as the signal statistics are known.
We use a linear predictor to infer $X_{M,t}$ from the latest received signals $X_{1,t-\Delta_1(t)},\ldots,X_{M,t-\Delta_M(t)}$, and use mean squared error (MSE) as the inference loss.
For each AoI vector, we compute the expected inference error from the target variance and the covariances among the target and received signals~\cite[Chapter~2.5.1]{brockwell2002introduction}.
These statistics are obtained by solving the Yule-Walker equations of the VAR model~\cite[Chapter~2.1.4]{kilian2006new}.
The VAR coefficients are scaled to ensure that these equations have valid solutions.
}

\begin{figure}[t]
    \centering    
    \begin{subfigure}[t]{0.49\linewidth}
        \centering
        \includegraphics[width=\linewidth]{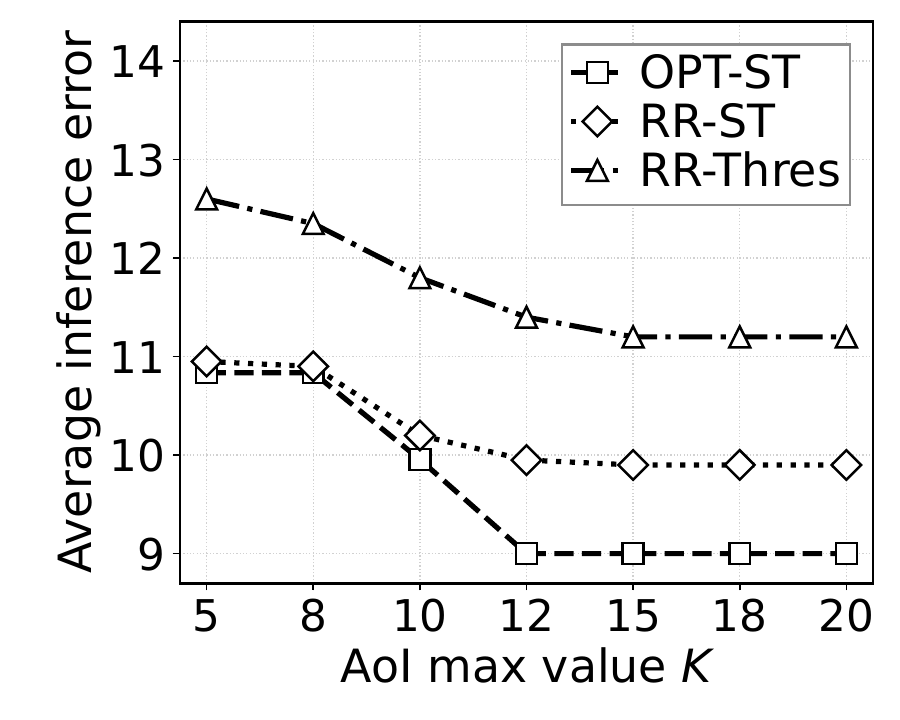}
        \caption{Average inference error.}
        \label{fig:k_error}
    \end{subfigure}\hfill
    \begin{subfigure}[t]{0.49\linewidth}
        \centering
        \includegraphics[width=\linewidth]{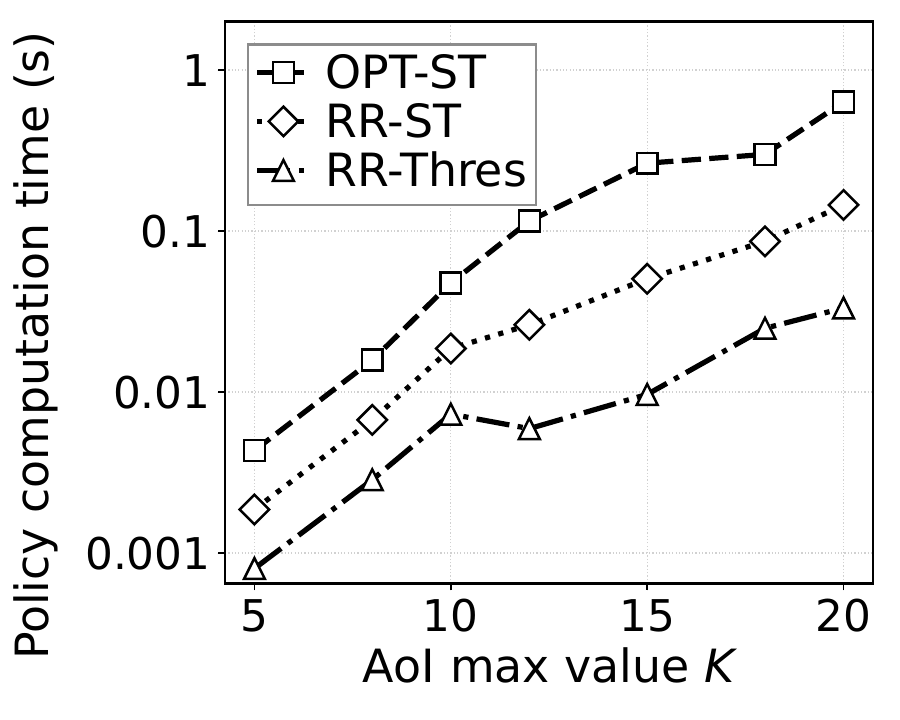}
        \caption{Policy computation time.}
        \label{fig:k_time}
    \end{subfigure}
    \caption{Average inference error and policy computation time under
    different AoI maximum values for $M=5$.}
    \label{fig:k_comparison}
\end{figure}


\subsection{The Impact of AoI on Inference Error} \label{sec:impact-of-AoI}
Fig.~\ref{fig:inference error} illustrates how the expected inference error varies with the AoI in the three cases. In Fig.~\ref{fig:2D}, the inference error increases faster with the AoI of modality 2 (LiDAR) than with that of modality 1 (control signal), indicating that LiDAR is more strongly correlated with the target signal.

The figures also show that the inference error is not necessarily monotonic in the AoI. For example, in Fig.~\ref{fig:3D}, when the AoIs of eye gaze and hand motion are fixed at 30, the inference error decreases as the RGB-video AoI approaches 30. This indicates that the inference error is smaller when the modalities have similar AoIs in this scenario.
Similarly, in Fig.~\ref{fig:5D}, the inference error is lowest when the AoI vector equals $(6,7,5,9,8)$. This is consistent with Eq.~\eqref{eq:xMt}, since we set $\boldsymbol{\nu}=(6,7,5,9,8)$ and the target signal $X_{M,t}$ depends primarily on the corresponding delayed terms $X_{m,t-\nu_m}$.

Overall, these results show that the inference error can be a non-monotonic and non-additive function of the AoI vector.

\subsection{Evaluation of Scheduling Policies}
{
We evaluate the four scheduling algorithms developed in Sections~\ref{sec:policy} and~\ref{sec:policy-multi}: Index, OPT-ST, RR-ST, and RR-Thres, where the Index policy is applicable only to the two-modality case. We choose the following four policies as baselines:
\begin{enumerate}[label=(\roman*)]
    \item \emph{Round-robin}: Modalities are selected in a fixed cyclic  order. It is a classical AoI-centric baseline and is optimal for minimizing the sum of AoIs under a symmetric setting~\cite{kadota2018scheduling}.
    \item \emph{Uniform random}: One modality is selected uniformly at random at each decision epoch.
    \item \emph{Greedy}: The modality that minimizes the inference error over the next transmission is selected.
    \item \emph{RL}: We implement a deep Q-network (DQN) with two fully connected hidden layers, each containing 128 ReLU units. The network is trained using Adam with a learning rate of $10^{-3}$. We use $\epsilon$-greedy exploration, with $\epsilon$ decreasing from $1$ to $0.05$. After training, the DQN yields a deterministic \emph{reformulated policy}, which is stored as a state-to-decision map.
\end{enumerate}

We use the inference error function characterized in Section~\ref{sec:impact-of-AoI} for evaluation. We set the evaluation horizon to $T=10{,}000$ time slots and $T_m^{\mathrm{pre}}=0$, independently sample each $T_m^{\mathrm{tx}}$ from 1 to 10 time slots, and randomly select the cyclic order for round-robin, RR-ST, and RR-Thres. Table~\ref{tab:cost_time_all} evaluates the policies from four aspects:
\begin{itemize}
    \item \textbf{Inference error}: the average expected inference error from $t=1$ to $T$.
    \item \textbf{Offline time}: the time required to compute the policy.
    \item \textbf{Storage overhead}: the memory required to store the resulting scheduling policy.
    \item \textbf{Online time}: the average time to compute the next scheduling decision during online inference.
\end{itemize}

Table~\ref{tab:cost_time_all} shows that OPT-ST achieves the minimum inference error in all three tasks. For $M=5$, RR-ST and RR-Thres reduce the offline computation time by approximately $5.2\times$ and $89.8\times$, respectively, compared with OPT-ST, while increasing the MSE from $8.08$ to $9.96$ and $11.2$. This shows their different trade-offs among inference error, offline computation time, and policy storage. Our algorithms also have low storage and online computation overhead. Even for $M=5$, OPT-ST requires only $2.44$~MiB to store the state-to-decision map, and its online decision time is only $4.20~\mu\text{s}$ because each decision only requires a lookup from this stored map.

RL achieves similar inference error to OPT-ST for $M=2$ and $M=3$, but requires longer offline computation because it relies on randomly sampled transitions during training. For $M=5$, OPT-ST achieves $20.0\%$ lower MSE than RL because the neural approximation of the Q-function does not guarantee convergence to the globally optimal policy.

}

{
\textbf{Impact of the AoI Maximum Value $\boldsymbol{K}$.}
To evaluate the impact of $K$, we set $K_m=K$ for all modalities and vary $K$ in Fig.~\ref{fig:k_comparison}. Fig.~\ref{fig:k_error} shows that the average inference errors decrease with $K$ and converge at $K=15$. Note that the converged values differ from those in Table~\ref{tab:cost_time_all} because this experiment uses a different simulation instance. Fig.~\ref{fig:k_time} shows that the computation times of OPT-ST and RR-ST increase more rapidly with $K$, while that of RR-Thres increases more slowly, consistent with our complexity analysis. At $K=15$, OPT-ST requires approximately $0.25$~s. These results show that an appropriate $K$ can significantly reduce computation time while maintaining similar inference performance. How to select $\boldsymbol{K}$ remains an interesting direction for future work.
}

\begin{table}[t]
\centering
\caption{
Comparison of approximation algorithms when the correlation parameter is $\theta = 0.9$.
}
\label{tab:approximation_comparison}

\renewcommand{\arraystretch}{1.08}

\begin{tabular}{lcccc}
\hline
Algorithm
& \shortstack{Average\\error}
& \shortstack{Offline\\time (s)}
& \shortstack{Storage\\overhead}
& \shortstack{Online time\\($\mu$s)}
\\
\hline

Round-robin
& 0.111
& 0
& 0 B
& 0.271
\\

Uniform random
& 0.124
& 0
& 0 B
& 7.53
\\

Greedy
& 0.132
& 0
& 0 B
& 19.2
\\

RL
& 0.118
& 29.6
& 642 B
& 2.09
\\
\hline

MGF-Corr
& 0.863
& 0.684
& 600 B
& 14.6
\\
\hline

OPT-ST (ours)
& \textbf{0.104}
& 0.384
& 300 B
& 2.35
\\

RR-ST (ours)
& 0.106
& 0.152
& 153 B
& 2.22
\\

RR-Thres (ours)
& 0.106
& 0.057
& 6 B
& 0.484
\\

\hline
\end{tabular}
\end{table}

\begin{figure}[t]
    \centering
    \includegraphics[width=0.5\columnwidth]{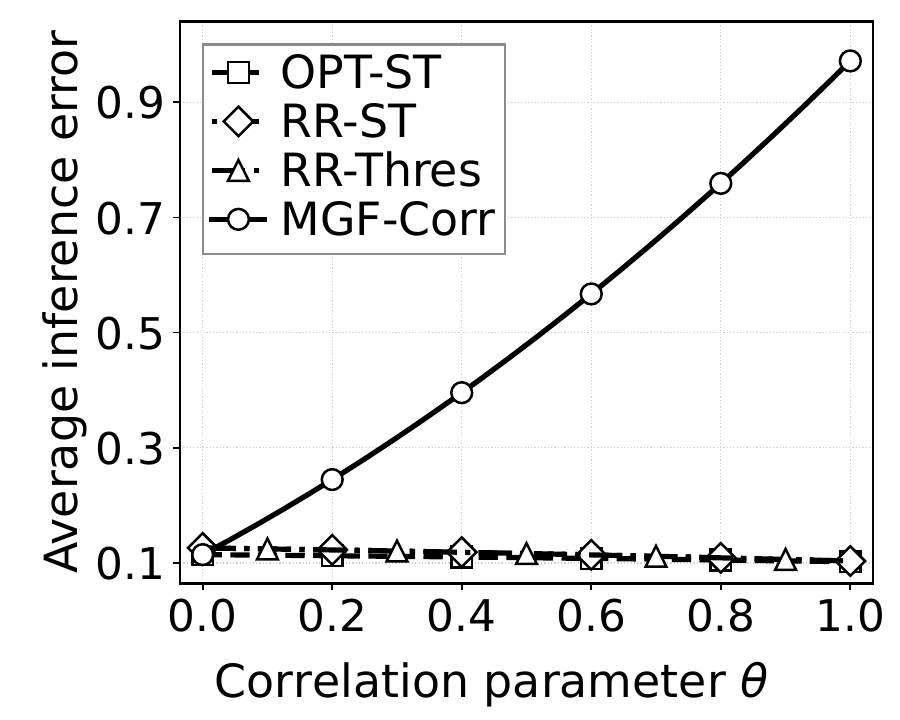}
    \caption{Average inference error across correlation levels.}
    \label{fig:correlation_comparison}
\end{figure}

\begin{table}[t]
\centering
\caption{Average inference error on the testing dataset.}
\label{tab:testing_results}
\begin{tabular}{lcc}
\hline
Algorithm & $M=2$ & $M=3$ \\
\hline
Round-robin        & 11.5         & 3.31          \\
Uniform random     & 7.59         & 5.59          \\
Greedy             & 9.04         & 4.66          \\
RL                 & 4.67         & \textbf{1.74} \\
\hline
Index (ours)       & \textbf{4.59} & --             \\
OPT-ST (ours)        & \textbf{4.59} & \textbf{1.74} \\
RR-ST (ours)         & \textbf{4.59} & 2.90          \\
RR-Thres (ours)          & \textbf{4.59} & 2.90          \\
\hline
\end{tabular}
\end{table}

\begin{figure}[t]
    \centering
    \includegraphics[width=0.50\columnwidth]
    {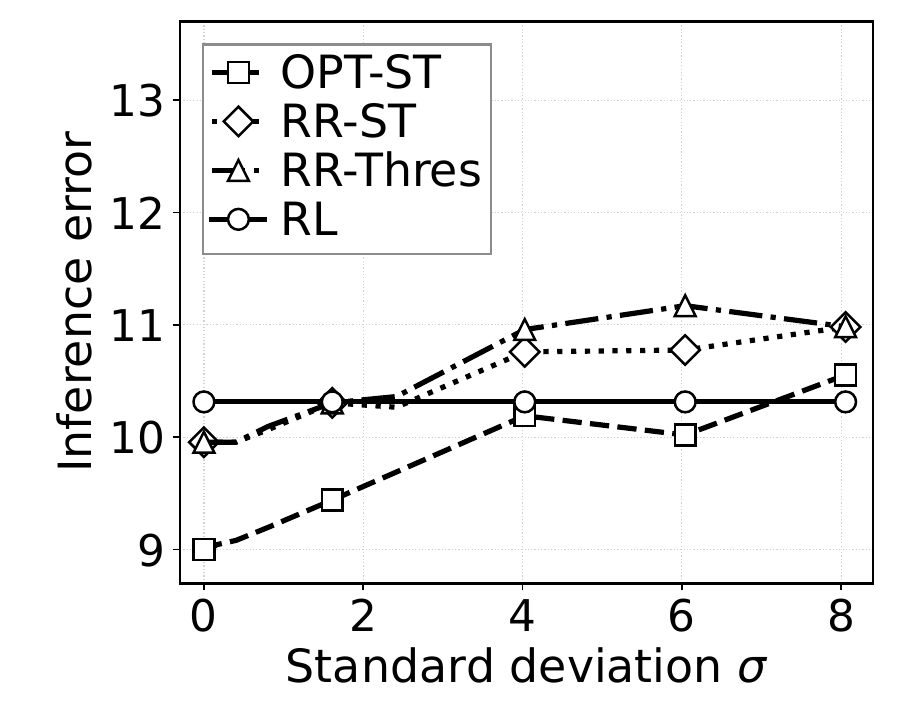}
    \caption{Impact of perturbations to the inference error function.}
    \label{fig:m5_robustness}
\end{figure}

{
\textbf{Comparison with Approximation Algorithm.}
In~\cite{shisher2025aoi}, an approximation-based algorithm, MGF-Corr, is proposed for correlated sources by approximating the general AoI-vector function with a weighted sum of scalar AoI functions. As MGF-Corr assumes unit transmission time, we compare it separately here. We consider a three-modality setting where modalities 1 and 2 are correlated with correlation controlled by $\theta$, while modality 3 is independent. Details of this setting are provided in Appendix~\ref{app:experimental-setup}.

Table~\ref{tab:approximation_comparison} shows the comparison of different policies under high correlation ($\theta=0.9$). MGF-Corr has an inference error of $0.863$, which is $8.3$ times that of OPT-ST and higher than those of all the other policies. This indicates that MGF-Corr can have a large approximation error under high correlation in this simulation setting. To better visualize the impact of correlation, we compare our algorithms with MGF-Corr under different values of $\theta$ in Fig.~\ref{fig:correlation_comparison}. When $\theta$ is small, MGF-Corr achieves performance close to that of our algorithms. As $\theta$ increases, however, MGF-Corr incurs increasingly higher inference error than our algorithms. This suggests that, while MGF-Corr performs comparably at low correlation, our algorithms better handle the complex AoI-vector function at high correlation in this simulation setting.


}

{
\textbf{Evaluation on Testing Data.}
For $M=2,3$, our previous analysis uses the inference error function obtained from the validation dataset.
We are interested in how our policies perform on the test dataset.
To this end, we average the test-sample losses for each AoI vector to obtain a test inference error function.
We then run our policies (obtained using the validation-set inference error function) on this new function.
Table~\ref{tab:testing_results} shows that all proposed policies achieve the lowest inference error for $M=2$; for $M=3$, OPT-ST matches RL for the lowest error, while RR-ST and RR-Thres outperform the remaining baselines.
These results show that our approach still maintains good performance on the test dataset.

To assess our policies' sensitivity to inaccuracies in $L$, we conduct a perturbation experiment for $M=5$.
Specifically, we add zero-mean Gaussian noise with variance $\sigma^2$ to each entry of $L$.
We compute our policies using the perturbed $L$ and evaluate them using the original $L$.
We run RL on the original $L$ to obtain the baseline policy.
Fig.~\ref{fig:m5_robustness} shows the results across $\sigma$, representing different accuracy levels.
From the figure, we see that small perturbations ($\sigma < 2$) have little effect on our policies, but performance degrades as perturbations increase.
One direction for future work is to adapt the scheduling policy when distribution shifts render $L$ inaccurate.
}

\section{Conclusion}

We studied scheduling for multimodal remote inference under limited network resources, with inference error modeled as a general function of the AoI vector.
We reduced the state space without losing optimality and showed that the reformulated SMDP was unichain for two modalities but could be multichain for more than two modalities.
We developed an optimal index-based threshold policy for two modalities, the switching-type policy OPT-ST for the general case, and two low-complexity algorithms, RR-ST and RR-Thres.
Numerical results showed that OPT-ST achieved the lowest inference error among the evaluated policies with less offline computation time than RL.
Our low-complexity algorithms, RR-ST and RR-Thres, traded higher inference error for lower offline computation time.

Future work could extend the current model to unreliable channels and multi-channel systems.
The resulting SMDPs would involve more complex decision sets and transition probabilities, calling for new analysis and algorithm design.
Reinforcement learning may help develop scheduling policies for these settings~\cite{yin2020application}.

\bibliographystyle{IEEEtran}
\begingroup
\microtypecontext{expansion=bibliography}
\bibliography{ieee}
\endgroup



\appendices

\section{Proof of Lemma~\ref{lem:existence}} \label{app:existence}
We apply results from SMDP theory and show that our problem satisfies sufficient conditions for the existence of an optimal stationary policy.
We first introduce the time average cost criterion for a general SMDP.
Let $N_t$ be the random number of decision epochs during the interval $[0, t]$.
The time average cost criterion $w_\pi(s)$ is given by the formula
\begin{equation} \label{eq:general-criterion}
\begin{aligned}
      w_\pi(s) \coloneq \limsup_{t \to \infty} \frac{1}{t} \mathbb{E}_\pi \Big [\sum_{i=0}^{N_t-1} & c(s_i, d_i, \tau_i) \\
      & + c(s_{N_t}, d_{N_t}, t - \sum_{i=0}^{N_t - 1}\tau_i) \Big ].   
\end{aligned}
\end{equation}
For completeness, we restate~\cite[Theorem 3]{yushkevich1982semi} as below.
\begin{theorem}[Restatement of Theorem 3~\cite{yushkevich1982semi}] \label{thm:restate}
Let $\mathbf{1}\{\cdot\}$ be an indicator function and $\tau$ be a random variable.
An SMDP with a finite state set $\mathcal{S}$ and a finite decision set $\mathcal{D}(s)$ has an optimal stationary policy under criterion $w$ if there exist positive constants $\varepsilon$ and $\eta$ and a positive decreasing function $\phi(t)$ with $\phi(\infty) = 0$ such that, for all $s \in \mathcal{S}$, $d \in \mathcal{D}(s)$, and $t \ge 0$,
\begin{enumerate}[label=(\roman*)]
    \item $\mathbb{P}\{\tau \ge \varepsilon | s, d\} > \eta$,
    \item $\mathbb{E} [\tau \mathbf{1}\{\tau \ge t\}] \le \phi(t)$,
    \item $\mathbb{E}[|c(s, d, \tau)|\mathbf{1}\{\tau \ge t\}] \le \phi(t)$,
    \item $|c(s, d, t)|\mathbb{P}\{\tau \ge t\} \le \phi(t)$.
\end{enumerate}
\end{theorem}
It is clear that our original problem formulation has a finite state set and a decision set, and the criterion $\bar{L}$ defined in Eq.~\eqref{eq:criterion} satisfies the average cost criterion in Eq.~\eqref{eq:general-criterion}. So, we only need to verify the four conditions in Theorem~\ref{thm:restate}.

First, the transition time is given by $T^{\mathrm{tx}}_{d_i}$, where $d_i$ denotes the action chosen at the $i$-th decision epoch.
As the transmission time is positive and discrete, we have $\mathbb{P}\{\tau \ge 1 \mid s, d\} = 1$ for all $s$ and $d$; hence, condition~(\romannumeral 1) holds.

The transmission time is also finite, so there exists $\tau_{\mathrm{max}}$ such that $\mathbb{P}\{ \tau > \tau_{\mathrm{max}}\} = 0$; therefore, we have 1) $\mathbb{E}\{\tau \mathbf{1}\{\tau \ge t\}\} \le \mathbb{E}\{\tau\} \le \tau_\mathrm{max}$ and 2) $\mathbb{E}[\tau \mathbf{1}\{\tau > \tau_{\mathrm{max}}\}] = 0$. Hence, condition (\romannumeral 2) holds if $\phi(t) \ge \tau_{\mathrm{max}}$ when $t \le \tau_{\mathrm{max}}$ and $\phi(t) > 0$ when $t > \tau_{\mathrm{max}}$.

For conditions~(\romannumeral 3) and~(\romannumeral 4), we use the fact that $|c(s, d, \tau)|$ is uniformly bounded because the expected inference error function $L$ is uniformly bounded (Assumption~\ref{ass:bounded}) and the transmission time is finite.
Therefore, it suffices to show that $\mathbb{E}[c_{\mathrm{max}}\mathbf{1}\{\tau \ge t\}] \le \phi(t)$ and $c_{\mathrm{max}}\mathbb{P}\{\tau \ge t\} \le \phi(t)$ hold for a sufficiently large positive constant $c_{\mathrm{max}}$. Thus, conditions~(\romannumeral 3) and~(\romannumeral 4) hold if $\phi(t) \ge c_\mathrm{max}$ when $t \le \tau_{\mathrm{max}}$ and $\phi(t) > 0$ when $t > \tau_{\mathrm{max}}$. 
Therefore, by defining
\[
\phi(t) = \max\{c_{\mathrm{max}}, \tau_{\mathrm{max}}\}
 e^{\tau_{\mathrm{max}}-t}, \qquad t\ge0,
\]
conditions~(\romannumeral 2), (\romannumeral 3), and~(\romannumeral 4) hold.\hfill\IEEEQED
\section{Proof of Lemma~\ref{lem:condition}} \label{app:condition}

The proof proceeds in two steps.
First, we characterize the optimal objective value.
Second, we show that an optimal policy always selects a fixed modality that attains this value.

Because Condition~\ref{con:reduction} does not hold, every optimal stationary policy switches modalities only finitely many times from every initial AoI state. Therefore, for every initial AoI state, there exist a finite transmission index $i_0$ and a modality $m\in\mathcal{M}$ such that the policy selects modality $m$ for every transmission after $i_0$.

Once the policy continuously selects modality $m$, the AoI of modality $m$ is reset to $T_m^{\mathrm{pre}}+T_m^{\mathrm{tx}}$ at each transmission completion, while the AoIs of all other modalities reach their respective maximum values after finitely many time slots. Consequently, the system repeatedly returns to the following AoI state at each transmission completion:
\begin{equation}
\left(T_m^{\mathrm{pre}}+T_m^{\mathrm{tx}};\boldsymbol{K}_{-m}\right)
\end{equation}
and the corresponding long-run average cost is
\begin{equation}
\frac{1}{T^{\mathrm{tx}}_m}
\sum_{t=0}^{T^{\mathrm{tx}}_m-1}
L\left(
T_m^{\mathrm{pre}}+T_m^{\mathrm{tx}}+t;
\boldsymbol{K}_{-m}
\right).
\label{eq:fixed-modality-cost}
\end{equation}

Because $L$ is uniformly bounded, the finite transient preceding this periodic evolution does not affect the long-run average cost. Hence, for any optimal stationary policy $\pi^*$ and any initial AoI state $\boldsymbol{\delta}$, its average cost is equal to the expression in \eqref{eq:fixed-modality-cost} for some $m\in\mathcal{M}$. Since every fixed-modality policy is feasible, optimality implies that this attained cost must be the minimum over all such policies. Therefore,
\begin{equation}\label{eq:optimal-lower-bound-fixed}
\bar{L}_{\pi^*}(\boldsymbol{\delta}) = 
\min_{m\in\mathcal{M}}
\frac{1}{T^{\mathrm{tx}}_m}
\sum_{t=0}^{T^{\mathrm{tx}}_m-1}
L\left(
T_m^{\mathrm{pre}}+T_m^{\mathrm{tx}}+t;
\boldsymbol{K}_{-m}
\right).
\end{equation}

The second step is to identify an optimal policy. Let
\begin{equation}
    m^*
    \in
    \arg\min_{m\in\mathcal{M}}
    \frac{1}{T^{\mathrm{tx}}_m}
    \sum_{t=0}^{T^{\mathrm{tx}}_m-1}
    L\left(
        T_m^{\mathrm{pre}}+T_m^{\mathrm{tx}}+t;
        \boldsymbol{K}_{-m}
    \right).
\end{equation}
By the preceding argument, from any initial AoI state, the policy that always selects modality $m^*$ reaches state
\begin{equation}
    \left(T_{m^*}^{\mathrm{pre}}+T_{m^*}^{\mathrm{tx}};\boldsymbol{K}_{-m^*}\right)
\end{equation}
after finitely many transmissions. Thus, it attains the optimal objective value in Eq.~\eqref{eq:optimal-lower-bound-fixed} from every initial AoI state. \hfill\IEEEQED

\section{Proof of Theorem~\ref{thm:equivalence}} \label{app:equivalence}

Assume Condition~\ref{con:reduction} holds, and fix an optimal stationary policy $\pi^*$ and an initial state $\boldsymbol{\delta}_0$ satisfying this condition.
The proof proceeds in three steps to establish the two assertions of Theorem~\ref{thm:equivalence}.
Steps~(\romannumeral 1) and~(\romannumeral 2) establish the existence of an optimal stationary reformulated policy $\pi_{\mathrm{re}}^* \in \Pi_{\mathrm{re}}^{\mathrm{S}}$ with the same average cost from every switching state, proving Eq.~\eqref{eq:reformulated-optimal-value}.
Using this policy, Step~(\romannumeral 3) shows that the same value is optimal for the original problem from every initial AoI state, proving Eq.~\eqref{eq:equivalence}.
Together, these steps yield
\begin{equation}
    \bar{L}_{\pi_{\mathrm{re}}^*}
    (\boldsymbol{\delta}_{\mathrm{re}})
    =
    \bar{L}_{\pi^*}(\boldsymbol{\delta})
\end{equation}
for every $\boldsymbol{\delta} \in \mathcal{K}$ and every $\boldsymbol{\delta}_{\mathrm{re}} \in \mathcal{S}_{\mathrm{re}}$. The common value will be denoted by $\bar{L}_{\mathrm{opt,re}}$.
The three steps are as follows.

(\romannumeral 1) Suppose that, starting from some initial AoI state $\boldsymbol{\delta}_0$, policy $\pi^*$ switches modalities infinitely many times. We show that there exist a stationary policy $\hat{\pi}_{\mathrm{re}} \in \Pi_{\mathrm{re}}^{\mathrm{S}}$ for the reformulated problem and a switching state $\boldsymbol{\delta}_{\mathrm{re},0} \in \mathcal{S}_{\mathrm{re}}$ such that
\begin{equation}
    \bar{L}_{\hat{\pi}_{\mathrm{re}}}(\boldsymbol{\delta}_{\mathrm{re},0})
    =  \bar{L}_{\pi^*}(\boldsymbol{\delta}_0).
\end{equation}

\noindent
(\romannumeral 2)
We show that there exists an optimal stationary policy $\pi_{\mathrm{re}}^* \in \Pi_{\mathrm{re}}^{\mathrm{S}}$
for the reformulated problem whose average cost is the same for every switching state. That is,
\begin{equation}\label{eq:reformulated-common-cost}
    \bar{L}_{\pi_{\mathrm{re}}^*}
    (\boldsymbol{\delta}_{\mathrm{re}})
    =
    \bar{L}_{\pi_{\mathrm{re}}^*}
    (\boldsymbol{\delta}'_{\mathrm{re}})
\end{equation}
for all
$\boldsymbol{\delta}_{\mathrm{re}},
\boldsymbol{\delta}'_{\mathrm{re}}
\in \mathcal{S}_{\mathrm{re}}$.

(\romannumeral 3)
We show that, if $\pi^*$ satisfies Condition~\ref{con:reduction}, then there exists an optimal policy $\hat{\pi}^*$ for the original problem that switches modalities infinitely many times starting from every initial AoI state.

We now explain how these steps establish Eqs.~\eqref{eq:reformulated-optimal-value} and~\eqref{eq:equivalence}.
By Step~(\romannumeral 1), there exist a stationary reformulated policy $\hat{\pi}_{\mathrm{re}}$ and a switching state $\boldsymbol{\delta}_{\mathrm{re},0}$ such that
\begin{equation}
    \bar{L}_{\hat{\pi}_{\mathrm{re}}}
    (\boldsymbol{\delta}_{\mathrm{re},0})
    =
    \bar{L}_{\pi^*}(\boldsymbol{\delta}_0).
\end{equation}
Moreover, by Step~(\romannumeral 2), the average cost under
$\pi_{\mathrm{re}}^*$ is the same for every switching state. Since this policy is optimal, denoting its common average cost by $\bar{L}_{\mathrm{opt,re}}$ gives Eq.~\eqref{eq:reformulated-optimal-value}.
Therefore, for every $\boldsymbol{\delta}_{\mathrm{re}} \in \mathcal{S}_{\mathrm{re}}$,
\begin{equation}
    \bar{L}_{\pi_{\mathrm{re}}^*}
    (\boldsymbol{\delta}_{\mathrm{re}}) = \bar{L}_{\pi_{\mathrm{re}}^*}(\boldsymbol{\delta}_{\mathrm{re},0})
    \le
    \bar{L}_{\hat{\pi}_{\mathrm{re}}}
    (\boldsymbol{\delta}_{\mathrm{re},0})
    = \bar{L}_{\pi^*}(\boldsymbol{\delta}_0).
\end{equation}

Finally, Step~(\romannumeral 3) shows that $\hat{\pi}^*$ is optimal and attains the common reformulated average cost from every initial AoI state
$\boldsymbol{\delta} \in \mathcal{K}$.
Hence,
\begin{equation}
    \bar{L}_{\pi_{\mathrm{re}}^*}
    (\boldsymbol{\delta}_{\mathrm{re}})
    \le \bar{L}_{\hat{\pi}^*}(\boldsymbol{\delta}) = 
    \bar{L}_{\pi^*}(\boldsymbol{\delta})
\end{equation}
for every
$\boldsymbol{\delta} \in \mathcal{K}$
and every
$\boldsymbol{\delta}_{\mathrm{re}} \in \mathcal{S}_{\mathrm{re}}$.
Executing $\pi_{\mathrm{re}}^*$ after reaching any switching state is feasible in the original problem and attains the same common average cost. Since $\pi^*$ is optimal, we have
\begin{equation*}
    \bar{L}_{\pi^*}(\boldsymbol{\delta}) \le \bar{L}_{\pi_{\mathrm{re}}^*} (\boldsymbol{\delta}_{\mathrm{re}}),
\end{equation*}
and thus 
\begin{equation*}
    \bar{L}_{\pi^*}(\boldsymbol{\delta}) = \bar{L}_{\pi_{\mathrm{re}}^*} (\boldsymbol{\delta}_{\mathrm{re}}).
\end{equation*}
Since $\pi^*$ is optimal for the original problem, this equality establishes Eq.~\eqref{eq:equivalence}.

\textbf{Proof of Step (\romannumeral 1).} The idea is to construct policy $\hat{\pi}_{\mathrm{re}}$. Because both $\pi^*$ and the state transitions are deterministic, starting from $\boldsymbol{\delta}_0$, policy $\pi^*$ induces a unique sequence of modality selections. Since this sequence contains infinitely many switches, the first switching time occurs after finitely many transmissions. Let $\boldsymbol{\delta}_{\mathrm{re},0}$ denote the AoI state at this switching time.

Consider the modality selection sequence starting from $\boldsymbol{\delta}_{\mathrm{re},0}$ under policy $\pi^*$. Let the current modality be $n$, and let $d_0,d_1,\ldots$ denote the subsequent modality selections.
Define
\begin{equation}
    i_0 \coloneqq  \min\{i\geq 0:d_i\neq n\}.
\end{equation}
Since $\pi^*$ switches modalities infinitely many times, $i_0$ is finite. This means that policy $\pi^*$ transmits modality $n$ for $i_0$ additional times and then switches to modality $d_{i_0}$. Accordingly, we define
\begin{equation}
    \hat{\pi}_{\mathrm{re}}(\boldsymbol{\delta}_{\mathrm{re}, 0}) \coloneqq  (i_0,d_{i_0}).
\end{equation}
Repeating this construction, we obtain a policy for the reformulated problem.
For each switching state not visited by the trajectory, we assign an arbitrary feasible decision.
As $\pi^*$ is stationary and the state transitions are deterministic, the same switching state always yields the same reformulated decision; hence, $\hat{\pi}_\mathrm{re}$ is a well-defined stationary policy.

By construction, each decision under $\hat{\pi}_\mathrm{re}$ represents exactly the sequence of modality selections made by $\pi^*$.
Hence, we have
\begin{equation}
    \bar{L}_{\hat{\pi}_{\mathrm{re}}}(\boldsymbol{\delta}_{\mathrm{re},0})
    =  \bar{L}_{\pi^*}(\boldsymbol{\delta}_{\mathrm{re},0}).
\end{equation}
Moreover, starting from $\boldsymbol{\delta}_0$, policy $\pi^*$ reaches $\boldsymbol{\delta}_{\mathrm{re},0}$ after a finite number of transmissions.
Since all transmission times are finite and $L$ is uniformly bounded, this finite transient does not affect the long-run average cost. Thus,
\begin{equation}
    \bar{L}_{\pi^*}(\boldsymbol{\delta}_{\mathrm{re},0}) = \bar{L}_{\pi^*}(\boldsymbol{\delta}_0),
\end{equation}
and we are done.

\textbf{Proof of Step (\romannumeral 2).} We construct an optimal stationary reformulated policy from the policy $\hat{\pi}_{\mathrm{re}}$ in Step~(\romannumeral 1).
Since $\mathcal{S}_{\mathrm{re}}$ is finite and the policy and transitions are deterministic, the trajectory from $\boldsymbol{\delta}_{\mathrm{re},0}$ eventually reaches a recurrent cycle, denoted by $\mathcal{R}^*$.
Each decision selected by this fixed policy has a finite $d_\tau$, so every transition on the cycle has finite duration. By Step~(\romannumeral 1), the average cost on $\mathcal{R}^*$ equals $\bar{L}_{\pi^*}(\boldsymbol{\delta}_0)$.
We retain the decisions of $\hat{\pi}_{\mathrm{re}}$ on $\mathcal{R}^*$ and construct a stationary policy $\pi_{\mathrm{re}}^*$ that reaches $\mathcal{R}^*$ from every switching state.

To see this, let $\mathcal{R}^* = \{s_1, s_2, \dots, s_j\}$, where $j$ is finite since the state space is finite. 
Because $\mathcal{R}^*$ is a recurrent class, starting from state $s_1$, the process returns to $s_1$ in finitely many steps. 
Additionally, as the state transitions are deterministic, we expand the decisions along this cycle into the sequence of individual modality transmissions and denote their modality choices by $\{d_1, d_2, \dots, d_i\}$.

Starting from any state $s \notin \mathcal{R}^*$, consider repeating the transmission sequence $\{d_1, d_2, \dots, d_i\}$; we can show that $s$ will go to $s_1$ eventually.
If the process enters $\mathcal{R}^*$ before transmitting modality $d_i$, it is trivial that $s$ transitions into $\mathcal{R}^*$.
Therefore, it suffices to show that, after finitely many repetitions of the transmission sequence, the process enters state $s_1$ from any state $s$.
We divide the proof into two cases: (\romannumeral 1) if $m$ is selected in the transmission sequence and (\romannumeral 2) if $m$ is not selected.

If $m$ is selected, then according to the AoI evolution in Eq.~\eqref{eq:AoI-evolution}, the AoI of modality $m$ is reset to $T_m^{\mathrm{pre}}+T_m^{\mathrm{tx}}$, regardless of its 
initial value. Therefore, at the end of the transmission sequence, the AoI of modality $m$ equals that of $s_1$.

If modality $m$ is not selected during the transmission sequence, then its AoI at $s_1$ must equal its maximum value $K_m$; otherwise, the sequence cannot return the process from $s_1$ to $s_1$. Moreover, starting from any state $s$, after finitely many repetitions of the transmission sequence, the AoI of modality $m$ reaches $K_m$.

Therefore, starting from any state $s \notin \mathcal{R}^*$, we can find a path to $\mathcal{R}^*$.

The repeated transmission sequence contains switches, so grouping transmissions between successive switches gives a finite path in the reformulated state graph to $\mathcal{R}^*$.
For each state outside $\mathcal{R}^*$, fix the first decision along a shortest such path, and retain $\hat{\pi}_{\mathrm{re}}$ on $\mathcal{R}^*$. Each decision outside $\mathcal{R}^*$ strictly decreases the shortest-path distance to $\mathcal{R}^*$, so every state eventually enters it. Each selected decision has finite duration, so this transient takes finite time.
Thus, $\pi_{\mathrm{re}}^*$ has a single recurrent class and a common average cost~\cite[Theorem 1]{denardo1968multichain}:
\begin{equation}
    \bar{L}_{\pi_{\mathrm{re}}^*}(s)
    = \bar{L}_{\pi^*}(\boldsymbol{\delta}_0),
    \qquad \forall s\in\mathcal{S}_{\mathrm{re}}.
\end{equation}
The cost of the finite transient vanishes in the time average because $L$ is uniformly bounded.

To prove optimality, take any stationary reformulated policy $\rho_{\mathrm{re}}$ and any initial switching state $s$. Its trajectory eventually reaches a recurrent cycle, and its average cost $\bar{L}_{\rho_{\mathrm{re}}}(s)$ equals the cost per unit time on that cycle.
Expand the cycle into a finite sequence of individual transmissions and repeat this sequence in the original problem starting from $\boldsymbol{\delta}_0$.
The same AoI reset and truncation argument used above shows that, after finitely many repetitions, the AoI vector agrees with the cycle at the corresponding phase. Hence, this feasible original policy attains $\bar{L}_{\rho_{\mathrm{re}}}(s)$ from $\boldsymbol{\delta}_0$.
Since $\pi^*$ is optimal at $\boldsymbol{\delta}_0$, we obtain
\begin{equation}
    \bar{L}_{\pi_{\mathrm{re}}^*}(s)
    = \bar{L}_{\pi^*}(\boldsymbol{\delta}_0)
    \le \bar{L}_{\rho_{\mathrm{re}}}(s).
\end{equation}
As $\rho_{\mathrm{re}}$ and $s$ were arbitrary, $\pi_{\mathrm{re}}^*$ is optimal from every switching state, and its common average cost is $\bar{L}_{\mathrm{opt,re}}$. This proves Eq.~\eqref{eq:reformulated-common-cost}.

\textbf{Proof of Step (\romannumeral 3).}
Our goal is to show that if policy $\pi^*$ satisfies Condition~\ref{con:reduction}, then there exists another optimal policy $\hat{\pi}^*$ under which the scheduler switches modalities infinitely many times starting from every initial state.
Let $\boldsymbol{\delta}_0$ be an initial state from which policy $\pi^*$ switches modalities infinitely many times. By Steps (\romannumeral 1) and (\romannumeral 2), the optimal reformulated policy $\pi_{\mathrm{re}}^*$ has the same average cost from every state in $\mathcal{S}_{\mathrm{re}}$, and this cost equals that of $\pi^*$ starting from $\boldsymbol{\delta}_0$. Denote this common value by
\begin{equation}
    g^*
    \coloneqq
    \bar{L}_{\pi^*}(\boldsymbol{\delta}_0)
    =
    \bar{L}_{\pi_{\mathrm{re}}^*}(s_{\mathrm{re}}),
    \qquad
    \forall s_{\mathrm{re}} \in \mathcal{S}_{\mathrm{re}}.
\end{equation}

We construct a policy $\hat{\pi}^*$ for the original problem as follows. If the initial state lies outside $\mathcal{S}_{\mathrm{re}}$, it first transmits two distinct modalities consecutively to enter $\mathcal{S}_{\mathrm{re}}$; otherwise, this initial step is omitted. It then follows the modality-selection sequence induced by $\pi_{\mathrm{re}}^*$. First, the scheduler switches modalities infinitely many times under policy $\hat{\pi}^*$. Moreover, the initial two transmissions, when needed, form only a finite transient. Since $L$ is uniformly bounded, this transient does not affect the long-run average cost. Therefore,
\begin{equation}
    \bar{L}_{\hat{\pi}^*}(s) = g^*,
\end{equation}
for any state $s$. 

We now show that $\hat{\pi}^*$ is optimal. Suppose it is not optimal starting from some state $s$. As $\pi^*$ is optimal, we must have
\begin{equation}
    \bar{L}_{\pi^*}(s) < g^*.
\end{equation}

If $\pi^*$ switches modalities infinitely many times starting from $s$, the construction in Step~(\romannumeral 1) gives a reformulated policy with the same average cost. This policy would achieve an average cost strictly smaller than $g^*$, contradicting the optimality of $\pi_{\mathrm{re}}^*$.

Otherwise, $\pi^*$ switches modalities only finitely many times. By the proof of Lemma~\ref{lem:condition}, $\pi^*$ eventually selects a fixed modality $m$, denoted by $m_s$. Always transmitting modality $m_s$ from $\boldsymbol{\delta}_0$ would also achieve $\bar{L}_{\pi^*}(s)<g^*$, contradicting the optimality of $\pi^*$ at $\boldsymbol{\delta}_0$.
Because $s \in \mathcal{S}$ is arbitrary, $\hat{\pi}^*$ is optimal starting from $s$, and we are done.

\hfill\IEEEQED

\section{Proof of Theorem~\ref{thm:main}} \label{sec:proof-main}
We first prove assertions~(\romannumeral 2) and~(\romannumeral 3) under the assumption that assertion~(\romannumeral 1) holds. The proof of assertion~(\romannumeral 1) is deferred to Appendix~\ref{sec:prop-tau}.

According to Proposition~\ref{prop:bellman-2}, we have
\begin{equation}
    \tau_m^* = \tau_m(g), \quad m = 1,2,
\end{equation}
where \(\tau_m(\cdot)\) is defined in Eq.~\eqref{eq:optimal-policy-beta}, and \(g\) satisfies the Bellman optimality equation in Eq.~\eqref{eq:bellman-2}. By assertion~(\romannumeral 1), we have
\begin{equation}
    \tau_m(g) =  \min \left\{ \theta : \gamma_m(\theta) \ge g \right\},
\end{equation}
which proves assertion~(\romannumeral 2).

For assertion~(\romannumeral 3), we show that $g$ is the root of Eq.~\eqref{eq:Lopt} by constructing a solution to the Bellman optimality equation~\eqref{eq:bellman-2}.
Given any $\beta \in \mathbb{R}$, we substitute $g = \beta$ and $\tau_m(\beta)$ into the Bellman optimality equation~\eqref{eq:bellman-2} and obtain
\begin{equation} \label{eq:bellman-beta}
\begin{aligned}
    h(s_{2 \to 1}) =   c_{2\to 1}(\tau_1(\beta)) - (\tau_1(\beta) T_1^{\mathrm{tx}} + T_2^{\mathrm{tx}})\beta + h(s_{1 \to 2}) , \\
    h(s_{1 \to 2}) =  c_{1 \to 2}(\tau_2(\beta)) - (T_1^{\mathrm{tx}} + \tau_2(\beta) T_2^{\mathrm{tx}})\beta + h(s_{2 \to 1}).
\end{aligned}
\end{equation}
By summing Eq.~\eqref{eq:bellman-beta} for each state, canceling $h(\cdot)$, and rearranging terms, we obtain an equation of $\beta$:
\begin{equation}\label{eq:beta}
\begin{aligned}
    c_{2\to 1}(\tau_1(\beta)) + c_{1 \to 2}(\tau_2(\beta))
    &- \beta \cdot \bigl((\tau_1(\beta) + 1)T_1^{\mathrm{tx}} \\
    &\qquad + (\tau_2(\beta) + 1)T_2^{\mathrm{tx}}\bigr) = 0,
\end{aligned}
\end{equation}
which is exactly Eq.~\eqref{eq:Lopt} in Theorem~\ref{thm:main}. Finally, because the inference error function is uniformly bounded, the AoI is finitely truncated, and the decision set is finite, both \(g\) and \(h(\cdot)\) are finite. Therefore, we prove that \(g\) is a solution to the Bellman optimality equation. According to~\cite[Theorem 3.1]{schweitzer1978functional}, it is unique.
\hfill\IEEEQED

\section{Proof of Assertion~(\romannumeral 1) in Theorem~\ref{thm:main}} \label{sec:prop-tau}
It suffices to prove the result for modality~1, as the proof for modality~2 is identical.
The finite decision ranges preserve the optimal value under Condition~\ref{con:reduction}: after at most $K-1$ additional transmissions of one modality, all other AoIs saturate, and transmitting that modality again would prevent any later switch under a deterministic stationary original policy. The finite ranges and the choice of the smallest minimizer ensure that $\tau_m(\beta)$ is well defined for every $\beta\in\mathbb{R}$. If $K=1$, the result follows immediately from $\gamma_m(0)=+\infty$. We therefore assume $K\ge2$ below.

We proceed by induction.
That is, we aim to show that for every integer $0 \le i \le K-1$, if $\min \{\theta : \gamma_1(\theta) \ge \beta\} = i$, then $\tau_{1}(\beta) = i$.

First, we prove the base case when $i= 0$. We have
\begin{equation}
    i = \min \{\theta : \gamma_1(\theta) \ge \beta\} = 0.
\end{equation}
Thus, we have $\gamma_1(0) \ge \beta$. From the definition of $\gamma_1$ in Eq.~\eqref{eq:index}, we obtain
\begin{equation*}
    \gamma_1(0) = \min_{k \in \{1,2,\dots,K-1\}} \frac{c_{2\to 1}(k) - c_{2\to 1}(0)}{kT_1^{\mathrm{tx}}} \ge \beta.
\end{equation*}
By rearranging terms, the inequality becomes
\begin{equation*}
    \min_{k \in \{1,2,\dots, K-1\}} \left [ \frac{c_{2\to 1}(k) - c_{2\to 1}(0)}{kT_1^{\mathrm{tx}}} - \beta \right] \ge 0.
\end{equation*}
Because $kT_1^{\mathrm{tx}} > 0$, multiplying both sides by $kT_1^{\mathrm{tx}}$ yields
\begin{equation*}
   \min_{k \in \{1,2,\dots, K-1\}} \left [c_{2\to 1}(k) - c_{2\to 1}(0) - kT_1^{\mathrm{tx}}\beta \right] \ge 0.
\end{equation*}
As $c_{2\to 1}(0)$ is independent of $k$, pulling it out yields
\begin{equation} \label{eq:base-case}
    \min_{k \in \{1,2,\dots, K-1\}} \left [c_{2\to 1}(k) - kT_1^{\mathrm{tx}}\beta \right] \ge c_{2\to 1}(0).
\end{equation}
The left-hand side (LHS) of Eq.~\eqref{eq:base-case} is the minimum of Eq.~\eqref{eq:optimal-policy-beta} over $\tau_1 \in \{1,2,\dots, K-1\}$;
and its right-hand side (RHS) is the objective value of Eq.~\eqref{eq:optimal-policy-beta} when $\tau_1 = 0$. 
Hence, Eq.~\eqref{eq:base-case} implies that $0$ is a minimizer. Since $\tau_1(\beta)$ is defined as the smallest minimizer, $\tau_{1}(\beta)=0$ if $i=0$.

For any integer $1\le j\le K-1$, assume that the result holds for $i = 0, 1, \dots, j - 1$.
We now show that $\tau_{1}(\beta) = j$ if
\begin{equation}\label{eq:induction-case}
    \min\{\theta : \gamma_1(\theta) \ge \beta\} = j.
\end{equation}
It is equivalent to the following two conditions: 
\begin{enumerate}[label=(\roman*)]
    \item $\gamma_1(i) < \beta$ for $i = 0, 1, \dots, j-1$;
    \item $\gamma_1(j) \ge \beta$.
\end{enumerate}

From condition (\romannumeral 1), we aim to show that $\tau_{1}(\beta) \ge j$ by contradiction.
Suppose $\tau_{1}(\beta) = i$ for some $i < j$.
Then, according to Eq.~\eqref{eq:optimal-policy-beta}, we have
\begin{equation}
    c_{2 \to 1}(i) - i T_1^{\mathrm{tx}} \beta \le c_{2\to1}(k) - k T_1^{\mathrm{tx}} \beta,
\end{equation}
for all $k\in\{i+1,\ldots,K-1\}$.
Rearranging terms, we obtain
\begin{equation}
    \frac{c_{2\to 1}(k) - c_{2\to 1}(i)}{(k - i) T_1^{\mathrm{tx}}} \ge \beta,
\end{equation}
for all $k\in\{i+1,\ldots,K-1\}$.
By definition,
\[
    \gamma_1(i) = \min_{i<k\le K-1}\frac{c_{2\to 1}(k) - c_{2\to 1}(i)}{(k - i) T_1^{\mathrm{tx}}}.
\]
Hence, $\gamma_1(i) \ge \beta$, contradicting condition (\romannumeral 1).

If $j=K-1$, the bound $\tau_1(\beta)\le K-1$ and the preceding inequality $\tau_1(\beta)\ge j$ give $\tau_1(\beta)=j$. It remains to consider $j<K-1$.

From condition (\romannumeral 2), we aim to show that $\tau_{1}(\beta) \le j$. From the definition of $\gamma_1$, we have
\begin{equation*}
     \gamma_1(j) = \min_{k \in \{1,2,\dots, K-1 - j\}} \frac{c_{2\to 1}(j + k) - c_{2\to 1}(j)}{kT_1^{\mathrm{tx}}} \ge \beta.
\end{equation*}
Multiplying both sides by $kT_1^{\mathrm{tx}}$ and rearranging terms yields
\begin{equation*}
    \min_{k \in \{1,2,\dots, K-1 - j\}} \left[c_{2\to 1}(j + k) - kT_1^{\mathrm{tx}} \beta\right] \ge c_{2\to 1}(j).
\end{equation*}
Subtracting $j T_1^{\mathrm{tx}}\beta$ from both sides yields
\begin{equation}\label{eq:condition-ii}
    \min_{k \in \{1,\dots, K-1 - j\}} [c_{2\to 1}(j + k) - (j + k)T_1^{\mathrm{tx}} \beta] \ge c_{2\to 1}(j) - j T_1^{\mathrm{tx}} \beta.
\end{equation}
Replacing $j + k$ with $\tau$, the LHS of Eq.~\eqref{eq:condition-ii} is equivalent to
\begin{equation*}
    \min_{\tau \in \{j + 1,j + 2,\dots, K-1\}} \left[c_{2\to 1}(\tau) - \tau T_1^{\mathrm{tx}} \beta\right].
\end{equation*}
The RHS of Eq.~\eqref{eq:condition-ii} is the objective value of Eq.~\eqref{eq:optimal-policy-beta} when $\tau = j$.
Hence, Eq.~\eqref{eq:condition-ii} shows that the objective value at $j$ is no larger than at any larger feasible decision. Since $\tau_1(\beta)$ is the smallest minimizer, it follows that $\tau_{1}(\beta)\le j$.
Combining $\tau_{1}(\beta) \ge j$ and $\tau_{1}(\beta) \le j$, we have $\tau_{1}(\beta) = j$. \hfill\IEEEQED

\section{Proof of Lemma~\ref{lem:chain_structure}}\label{app:chain}
We first prove the case when $M=2$. We use the result stated in~\cite[Exercise 4.3]{gallager2013stochastic}: \emph{If a finite-state Markov chain has a state that is accessible from every other state, then the Markov chain has exactly one recurrent class.}
Therefore, it suffices to show that, under every stationary policy, there exists a state that is accessible from every other state.
Note that a state \(j\) is said to be \emph{accessible} from a state \(i\) if, starting from \(i\), there is a positive probability of reaching \(j\) within a finite time.
We show that both $s_{2\to1}$ and $s_{1\to2}$ have this property.

For the reformulated problem, by the definition of switching states in Eq.~\eqref{eq:def-switch}, the system has only two switching states, namely $s_{2\to1}$ and $s_{1\to2}$. Under any stationary policy, since the number of consecutive transmissions is finite, the system alternates between them with probability \(1\).


Next, we present a counterexample showing that when $M > 2$, the SMDPs of the two problems can be multichain.
Let $\mathcal{M} = \{1,2,3\}$, with $T_m^{\mathrm{pre}}=0$ and $T_m^{\mathrm{tx}}=1$ for all $m\in\mathcal{M}$. Consider a stationary policy in $\Pi^\mathrm{S}$ that induces the following state transition:
\begin{enumerate}[label=(\roman*)]
    \item (1, 2, 3) $\xrightarrow{3}$ (2, 3, 1) $\xrightarrow{2}$ (3, 1, 2) $\xrightarrow{1}$ (1, 2, 3) $\cdots$,
    \item (1, 2, 4) $\xrightarrow{1}$ (1, 3, 5) $\xrightarrow{3}$ (2, 4, 1) $\xrightarrow{3}$ (3, 5, 1) $\xrightarrow{2}$ (4, 1, 2) $\xrightarrow{2}$ (5, 1, 3) $\xrightarrow{1}$ (1, 2, 4) $\cdots$,
\end{enumerate}
Each arrow is labeled with the selected modality. The transitions satisfy the AoI evolution in Eq.~\eqref{eq:AoI-evolution}, and the policy assigns a unique action to each listed state. Thus, this is a valid stationary policy.
The sets $\{(1,2,3),\allowbreak(2,3,1),\allowbreak(3,1,2)\}$ and $\{(1,2,4),\allowbreak(1,3,5),\allowbreak(2,4,1),\allowbreak(3,5,1),\allowbreak(4,1,2),\allowbreak(5,1,3)\}$ form two distinct recurrent classes. Hence, the policy has at least two recurrent classes.
By rewriting the states and decisions in terms of the reformulated problem, we obtain the following counterexample for the reformulated problem:\begin{enumerate}[label=(\roman*)]
    \item $(1,2,3) \xrightarrow{(0,3)} (2,3,1) \xrightarrow{(0,2)} (3,1,2) \xrightarrow{(0,1)} (1,2,3) \cdots,$
    \item $(1,2,4) \xrightarrow{(1,3)} (2,4,1) \xrightarrow{(1,2)} (4,1,2) \xrightarrow{(1,1)} (1,2,4) \cdots.$
\end{enumerate}
\hfill\IEEEQED

\section{Proof of Proposition~\ref{prop:foft-state}}\label{app:CD}
The key idea is to express the AoI at each switching time using the decision variable $\boldsymbol{\tau}$.
Fix a vector of extra consecutive transmissions
\(\boldsymbol{\tau}=(\tau_1,\ldots,\tau_M)\), and suppose the switching order is fixed as
\[
1 \to 2 \to \cdots \to M \to 1.
\]
We first consider $s_{M-1\to M}(\boldsymbol{\tau})$, the state reached when the system switches from modality $M-1$ to modality $M$.

At this switching instant, modality \(M\) has just been successfully
delivered, so its AoI is $\Delta_M = T_M^{\mathrm{pre}}+T_M^{\mathrm{tx}}$.
Moreover, modality \(M-1\) was last delivered immediately before the first
transmission of modality \(M\), and hence $\Delta_{M-1} = T_{M-1}^{\mathrm{pre}}+T_{M-1}^{\mathrm{tx}}+T_M^{\mathrm{tx}}$.

Next, for each \(m\in\{1,\ldots,M-2\}\), after the last successful
transmission of modality \(m\), the system serves modalities
\(m+1,m+2,\ldots,M-1\) in order, where each modality \(j\) is transmitted
\(\tau_j+1\) times, and then serves modality \(M\) once. Therefore, the AoI
of modality \(m\) at state \(s_{M-1\to M}(\boldsymbol{\tau})\) is
\begin{equation}
    \Delta_m
=
T_m^{\mathrm{pre}}+T_m^{\mathrm{tx}}+\sum_{j=m+1}^{M-1}(\tau_j+1)T_j^{\mathrm{tx}}+T_M^{\mathrm{tx}}.
\end{equation}
Therefore, we prove that the AoI of every modality is uniquely determined by \(\boldsymbol{\tau}\). It means that there is only one switching state from modality $M-1$ to $M$. The same argument applies to every switching state. Hence, the system cycles among $M$ states. We denote these states by
\begin{equation}
  s_{1\to2}(\boldsymbol{\tau}),\,s_{2\to3}(\boldsymbol{\tau}),\,\ldots,\,s_{M\to1}(\boldsymbol{\tau})
\end{equation}

Next, we compute the average cost under $\boldsymbol{\tau}$. Consider one complete cycle. Modality \(i\) is transmitted exactly \(\tau_i + 1\) times (once for switching and \(\tau_i\) additional times). So, the cycle duration is
\begin{equation}
    D(\boldsymbol{\tau})=\sum_{i=1}^M (\tau_i + 1) T_i^{\mathrm{tx}}.
\end{equation}
Hence, the average inference
error induced by \(\boldsymbol{\tau}\) is
\begin{equation}
\frac{\sum_{i=1}^M c_{\mathrm{re}}(s_{(i-1)\to i}(\boldsymbol{\tau}),\tau_i)}
     {\sum_{i=1}^M (\tau_i + 1) T_i^{\mathrm{tx}}},
\end{equation}
which leads to Eq.~\eqref{eq:foft-problem}.
\hfill\IEEEQED
\section{Experimental Setup}
\label{app:experimental-setup}

\subsection{Two-Modality Case: Simulated Robot State Prediction}
We generate time-series data using the Bipedal Walker simulation environment~\cite{towers2024gymnasium}, in which a four-joint robot traverses terrain containing obstacles. The robot is controlled by a trained TD3-FORK agent~\cite{9683288}. We record its velocity, joint control signals, and LiDAR range measurements. The prediction target is the robot's current velocity. Modality 1 consists of the control signals, and modality 2 consists of the LiDAR measurements.

The predictor is an LSTM network with an input layer, a hidden layer containing 20 LSTM cells, and a fully connected output layer. It uses the historical inputs corresponding to the two modality ages to predict the current velocity. We use mean squared error (MSE) as the inference loss.

\subsection{Three-Modality Case: Hand Pose Prediction}
We use the HoloAssist dataset~\cite{wang2023holoassist}, which contains synchronized multimodal streams collected by an augmented reality device while users perform physical manipulation tasks. The three modalities are RGB video, hand motion, and eye gaze. The target is the future sequence of hand joint positions.

We adopt a Seq2Seq predictor~\cite{sutskever2014sequence} based on the network implementation described in~\cite[Appendix S.5]{wang2023holoassist}. To incorporate data freshness, we concatenate each modality's AoI with its feature vector. The model takes a 3-second input clip, with the modality inputs shifted according to their respective ages, and predicts hand poses over the subsequent 1.5 seconds. The inference loss is the Euclidean distance between predicted and ground-truth hand joint positions, measured in centimeters.

\subsection{Five-Modality Case: Vector Autoregressive Process Prediction}
The model-based application uses the five-dimensional VAR process defined in Section~\ref{sec:numerical}. The first four modalities follow first-order autoregressive dynamics, while the fifth modality depends on delayed values of all five modalities, as specified in~\eqref{eq:xMt}. The model order is $p=10$. Before coefficient scaling, we set $\rho_m=0.95$ for $m=1,\ldots,4$ and $b_m=0.95$ for $m=1,\ldots,5$. Each lag $\nu_m$ is selected from $\{1,\ldots,10\}$, and each innovation has zero mean and standard deviation $0.005$. The lag vector used in Fig.~\ref{fig:5D} is $(6,7,5,9,8)$. The VAR coefficients are scaled to obtain valid stationary covariance statistics.

We infer $X_{5,t}$ from the latest received value of each modality using a linear predictor and compute its MSE following~\cite[Chapter~2.5.1]{brockwell2002introduction}. The required covariances are obtained by solving the VAR Yule--Walker equations~\cite[Chapter~2.1.4]{kilian2006new}.

\subsection{Experimental Setup for the Comparison with MGF-Corr}
We consider three zero-mean, unit-variance Gaussian AR(1) sources. We vary the cross-correlation coefficient $\rho\geq0$ between modalities 1 and 2, with larger values indicating stronger correlation. The cross-correlation coefficients between modality 3 and the other two modalities are zero. We construct an asymmetric prediction target
\begin{equation}
\label{eq:mgf-corr-target}
    Y_t=c_1(X_{1,t}+X_{2,t})^2+c_2X_{3,t},
\end{equation}
where $c_1=1/[2\sqrt{2}(1+\rho)]$ and $c_2=0.2$.
We use MSE as the loss function. The inference error function is obtained by analytically computing the minimum MSE for each AoI vector using Gaussian conditional moments.

\section{Derivation of the Reformulated Transition Cost}
\label{app:transition-cost}

{
Consider a switching time at which modality $n$ has just been delivered.
Let $s_{\mathrm{re}}=\boldsymbol{\delta}$ be the current AoI vector and let $d_{\mathrm{re}}=(d_\tau,d_s)$ be the decision.
Here, $\delta_n=T_n^{\mathrm{pre}}+T_n^{\mathrm{tx}}$.
The transition consists of $d_\tau$ further transmissions of modality $n$, followed by one transmission of modality $d_s$.
Throughout this derivation, AoI vectors are truncated componentwise at $\boldsymbol{K}$.

During the repeated transmissions of modality $n$, its AoI resets to $T_n^{\mathrm{pre}}+T_n^{\mathrm{tx}}$ at each delivery, while the AoIs of all other modalities continue to increase.
Index these transmissions by $k=0,\ldots,d_\tau-1$.
At elapsed time $kT_n^{\mathrm{tx}}+t$, where $0\le t<T_n^{\mathrm{tx}}$, the AoI vector is
\[
\left(T_n^{\mathrm{pre}}+T_n^{\mathrm{tx}}+t;\,
\boldsymbol{\delta}_{-n}+(kT_n^{\mathrm{tx}}+t)\mathbf{1}_{M-1}\right).
\]
Thus, the cumulative cost of these transmissions is
\begin{equation}\label{eq:cost-repeat}
\sum_{k=0}^{d_\tau-1}\sum_{t=0}^{T_n^{\mathrm{tx}}-1}
L\!\bigl(\delta_n+t;\boldsymbol{\delta}_{-n}+(kT_n^{\mathrm{tx}}+t)\mathbf{1}_{M-1}\bigr).
\end{equation}

The transmission of modality $d_s$ begins at elapsed time $d_\tau T_n^{\mathrm{tx}}$.
At that time, modality $n$ has AoI $T_n^{\mathrm{pre}}+T_n^{\mathrm{tx}}$, and every other modality has aged by $d_\tau T_n^{\mathrm{tx}}$ since the initial switching time.
All AoIs increase until this transmission completes.
Its cumulative cost is therefore
\begin{equation}\label{eq:cost-switch}
\sum_{t=0}^{T_{d_s}^{\mathrm{tx}}-1}
L\!\bigl(\delta_n+t;\boldsymbol{\delta}_{-n}+(d_\tau T_n^{\mathrm{tx}}+t)\mathbf{1}_{M-1}\bigr).
\end{equation}
The delivery of modality $d_s$ occurs at the next switching time and is excluded from the current transition's cost, consistent with the interval $[S_i,S_{i+1})$ used in the original problem.
Adding Eqs.~\eqref{eq:cost-repeat} and~\eqref{eq:cost-switch} gives Eq.~\eqref{eq:reformulated-cost}.
}

\section{Time Complexity Analysis}\label{app:complexity}

\textbf{Optimal switching-type policy (OPT-ST).}
Consider one iteration of Alg.~\ref{alg:mcpi}. In the policy-evaluation step, the recurrent classes can be identified in \(\mathcal{O}(|\mathcal{S}_{\mathrm{re}}|)\) time because the induced transition is deterministic. Then, computing \(g(\cdot)\) and \(h(\cdot)\) requires only a constant number of visits to each state. Therefore, the policy-evaluation step costs
\begin{equation}
\mathcal{O}\!\left(|\mathcal{S}_{\mathrm{re}}|\right).
\end{equation}

In the policy-improvement step, the algorithm checks every feasible state-decision pair a constant number of times. Since each state has exactly \(|\mathcal{D}_{\mathrm{re}}(s)|\) feasible decisions, this step costs
\begin{equation}
\mathcal{O}\!\left(|\mathcal{S}_{\mathrm{re}}||\mathcal{D}_{\mathrm{re}}(s)|\right).
\end{equation}

For OPT-ST, each of the $M(M-1)$ ordered switching pairs fixes two AoI values, leaving at most $K^{M-2}$ states per pair. Thus, $|\mathcal{S}_{\mathrm{re}}|\le M(M-1)K^{M-2}$. At each state, the decision chooses $d_\tau\in\{0,\ldots,K-1\}$ and $d_s\in\mathcal{M}\setminus\{n\}$, where $n$ is the current modality, giving $|\mathcal{D}_{\mathrm{re}}(s)|=(M-1)K$.

Hence, one iteration of MPI for OPT-ST costs
\begin{equation}
\mathcal{O}\!\left(M(M-1)^2K^{M-1}\right).
\end{equation}

\textbf{Round-robin-like switching-type policy (RR-ST).}
Under the fixed switching order, each modality has a fixed predecessor, leaving only $M$ switching pairs and at most $MK^{M-2}$ states. The next modality is fixed, so the decision reduces to $d_\tau\in\{0,\ldots,K-1\}$, giving $K$ decisions per state. Applying the same argument as above, one iteration costs
\begin{equation}
\mathcal{O}\!\left(MK^{M-2}\cdot K\right)=\mathcal{O}\!\left(MK^{M-1}\right).
\end{equation}

\textbf{Policy iteration for the original problem.}
The original problem has at most $K^M$ states and $M$ decisions per state. Applying the same argument again, one iteration costs
\begin{equation}
\mathcal{O}\!\left(K^M\cdot M\right)=\mathcal{O}\!\left(MK^M\right).
\end{equation}

\textbf{Round-robin-like threshold policy (RR-Thres).}
In Alg.~\ref{alg:foft-cd}, each coordinate update searches over at most \(K\) candidate values of one threshold. {Hence, sweeping over $M$ modalities requires $\mathcal{O}(MK)$ objective evaluations. When each objective evaluation is counted as one operation, the time complexity per sweep is}
\begin{equation}
\mathcal{O}(MK).
\end{equation}
\hfill\IEEEQED

\end{document}